\documentclass[11pt]{article}
\usepackage[margin=1in]{geometry}
\usepackage[T1]{fontenc}
\usepackage{lmodern,newunicodechar}
\newunicodechar{–}{\textendash}
\usepackage{amsmath,amssymb,amsthm,booktabs,graphicx,authblk,microtype,array}
\usepackage[numbers,sort&compress]{natbib}
\usepackage[dvipsnames]{xcolor}
\usepackage[colorlinks=true,linkcolor=MidnightBlue,citecolor=MidnightBlue,urlcolor=MidnightBlue]{hyperref}
\usepackage{placeins,caption,needspace,etoolbox}
\AtBeginEnvironment{table}{\small}
\pretocmd{\section}{\Needspace{6\baselineskip}}{}{}
\pretocmd{\subsection}{\Needspace{4\baselineskip}}{}{}

\newcommand{\sigmatf}{\sigma_{\mathrm{TF}}}

\newtheorem{proposition}{Proposition}
\AtBeginEnvironment{proposition}{\Needspace{5\baselineskip}}
\newcounter{algorithm}
\newenvironment{numericalalgorithm}[1]{%
  \refstepcounter{algorithm}\par\medskip\noindent
  \begin{minipage}{\linewidth}\hrule\vspace{5pt}
  \textbf{Algorithm \thealgorithm. #1}\par\vspace{3pt}\hrule\vspace{5pt}\small
}{\par\vspace{5pt}\hrule\end{minipage}\par\medskip}
\graphicspath{{figs/}}

\title{Forecasting Without Full Identification:\\Reliability and Computational Cost in Clot-Growth Simulation}
\author[1,2]{Lennon J. Shikhman}
\author[3]{Lily DiPaulo}
\author[4]{Ying Qian}
\author[4]{He Li\thanks{Corresponding author: \href{mailto:he.li3@uga.edu}{he.li3@uga.edu}}}
\affil[1]{College of Computing, Georgia Institute of Technology, 801 Atlantic Drive, Atlanta, GA 30332-0280, USA}
\affil[2]{Department of Mathematics and Systems Engineering, Florida Institute of Technology, 150 W. University Blvd., Melbourne, FL 32901, USA}
\affil[3]{Department of Mathematics, University of Pittsburgh, 301 Thackeray Hall, 139 University Place, Pittsburgh, PA 15260, USA}
\affil[4]{School of Chemical, Materials and Biomedical Engineering, University of Georgia, 220 Riverbend Road, Athens, GA 30602, USA}
\date{}
\begin{document}
\maketitle
\vspace{-2em}
\begin{abstract}
We present a framework for forecasting clot growth from sparse early
measurements with incomplete biochemical information. Using an archive of
1,000 simulated trajectories, we compare inverse search, which weights
candidate inputs by their agreement with early observations, with a direct
multilayer perceptron (MLP) that predicts future clot size from the available
measurements. Inverse search achieves higher accuracy when tissue factor
alone, or tissue factor and fibrinogen, are unknown. When all five inputs
are unknown, direct prediction achieves 2.16\% pooled relative $L_2$
forecast error, outperforming search with 8,192 candidates using either
a forward MLP (3.25\%) or a latent neural ordinary differential equation
model (3.36\%). These comparisons demonstrate how the available biochemical
information shapes the relative performance of direct and inverse
forecasting. To assess forecast reliability, we derive a conditional error
bound that separates candidate disagreement, weight on poor early fits,
surrogate error, and candidate coverage. Increasing randomized
quasi-Monte Carlo candidate budgets from 512 to 32,768 reduces numerical
variation. A covariance score ranks additional assays by their expected
forecast value while accounting for measurement noise. At 10\% range-scaled
assay noise, incorporating assay uncertainty reduces forecast error from
about 5\% to below 3\% for both surrogates. An independent reaction cascade
verifies candidate integration against tensor quadrature, with mean
normalized forecast-integration error below $10^{-5}$ at 32,768 candidates.
Together, these results demonstrate accurate clot-growth forecasting
without full biochemical identification and provide tools for selecting
prediction strategies and additional measurements according to accuracy,
reliability, and computational cost.
\end{abstract}
\section{Introduction}

A short observation record may support a forecast even when some inputs
to a dynamical system are unknown. One approach uses a forward surrogate,
a fast approximation of the simulator. Plausible inputs are sampled, their
trajectories are predicted, and candidates that fit the early observations
receive more weight. Forecast reliability then depends on the disagreement
among candidate futures, the surrogate's approximation error, and the
accuracy of the candidate integration. Concentrating the weights on a few
candidates does not establish that all three contributions are small.
Inverse-problem and identifiability theory distinguish information supplied
by the observations from assumptions supplied by the model and prior
\cite{stuart2010inverse,raue2009identifiability}.

Clot growth provides a useful setting for this problem. Platelet activation
and fibrin formation work together in blood clotting
\cite{versteeg2013new,furie2008mechanisms,palta2014overview}, with biochemical
feedback and thrombin regulation shaping the response
\cite{butenas2002blood,tanaka2009blood}. Experiments on hemophilia document
changes in coagulation and spatial clot growth
\cite{cawthern1998blood,ovanesov2002hemophilia}; experiments on combined
platelet stimuli show why inputs cannot always be considered separately
\cite{chatterjee2010pairwise}. The factors controlling the start of clotting
need not control its later propagation
\cite{panteleev2006spatial,ovanesov2005initiation,kuprash2018sensitivity}.
The spatial model underlying the present archive describes coagulation
initiated by tissue-factor-expressing cells in nonflowing plasma through
coupled reaction and diffusion processes \cite{panteleev2006spatial}.
That study distinguishes initiation near the activating surface from later
propagation sustained by the intrinsic pathway. It also describes how the
protein C pathway limits the spatial extent of clotting. A clot-size record
therefore combines the effects of several interacting processes. It does
not measure each biochemical state separately. This motivates our question:
how much can early growth reveal about later growth when only some
biochemical inputs are known?

Coagulation models represent these processes at different levels of detail
\cite{anand2022computational,leiderman2014overview}. Reaction models describe
activation thresholds, surface binding and thrombin production
\cite{beltrami1995mathematical,kuharsky2001surface,bungay2003mathematical,hockin2002model}.
Spatial models add transport and platelet deposition
\cite{zarnitsina1996mathematical,fogelson2005coagulation,leiderman2011grow,leiderman2013influence}.
Traveling-wave descriptions of thrombin activity \cite{dashkevich2012thrombin}
and models of clot initiation and growth under flow
\cite{bouchnita2017mathematical,bouchnita2019multiscale,bouchnita2020mathematical,bouchnita2023thrombin}
extend this setting. These models contain internal states that a single
clot-size curve cannot distinguish. Because many inputs are difficult to
measure, practical measurements can instead be used to infer them
\cite{albannoud2024biomarkers,albannoud2025kinetics}.
For forecasting, the question is whether this additional inference improves
accuracy enough to justify its computational cost.

This question leads to two uses of the same simulation archive. A forward
surrogate learns to predict a trajectory from complete biochemical inputs.
When some inputs are unknown, inverse search finds and weights candidates
that explain the observations. A direct predictor instead learns to predict
future clot size from the measurements already available. Inverse search
can also provide plausible biochemical explanations or help choose an
additional assay. A forecast alone may not require those outputs.

We compare the two strategies using the same early observations and three
patterns of missing inputs. When only tissue factor (TF) is unknown, all
four factor concentrations are supplied. The TF+Fg setting also withholds
fibrinogen (Fg). The all-five setting withholds every biochemical input.
These settings control the information available within one simulation
archive; they are not clinical assay protocols.

Surrogates and multifidelity methods already approximate expensive biological
models \cite{guerrero2023multifidelity,gherman2023bridging,qian2024coagulonet,albannoud2026surrogates}.
Examples include cerebral-circulation uncertainty quantification
\cite{yuhn2022uq}, cell-based simulation \cite{comlekoglu2025surrogate},
aortic-flow modelling \cite{daneker2024transfer} and mechanistic learning
from sparse biological measurements \cite{lagergren2020binn}.
Neural ordinary differential equations (ODEs), latent ODEs and conditioning
on arbitrary subsets of observations are established approaches
\cite{chen2018neuralode,rubanova2019latentode,ivanov2019arbitrary}.
These methods offer several ways to predict a trajectory. Their existence
does not tell us whether inverse search with an ODE surrogate, inverse search
with a feedforward network, or direct prediction works best for this task.
Each approach provides different information and has different approximation
errors and computational costs.

Our main empirical finding is that the relative accuracy of direct and
inverse forecasting reverses across the three information settings
(Figure~\ref{fig:direct-baseline}). Inverse search is more accurate
when tissue factor alone, or tissue factor and fibrinogen, are unknown;
direct prediction is more accurate when all five inputs are unknown,
with 2.16\% pooled relative $L_2$ error versus 3.25--3.36\% for search at
8,192 candidates. Thus the available biochemical information changes whether
the accuracy gain from inverse search justifies its additional cost. The
tested patterns vary both the number and identity of missing factors, so
the result does not identify a universal threshold for the number of unknown
inputs. Candidate-budget experiments and failure diagnostics examine the
reliability of the search. Noisy-assay experiments then assess the value of
additional biochemical information. A conditional error bound separates
disagreement about the future from surrogate and candidate error. A
covariance score applies established goal-oriented design principles
\cite{chaloner1995design,attia2018goal} to choose measurements that target
forecast accuracy. A reaction system with known equations lets us verify
the inference procedure against independent quadrature.

\section{Benchmarks and prediction tasks}
\label{mc:data}

\subsection{Archived clot-growth trajectories}
The archive contains 1,000 distinct input vectors
$\theta=(\sigmatf,\mathrm{Fg},\mathrm{IX},\mathrm{VIII},\mathrm{V})$
and their nonnegative, nondecreasing size curves $u(t;\theta)$. Each curve has
121 samples from 0.0008 to 59.9992 minutes. All entries are finite.
The trajectories were generated using the original code for the spatial
coagulation model of Panteleev et al.\ \cite{panteleev2006spatial}, as
confirmed by the corresponding author. Its first 243 inputs form a complete
three-level factorial design; the remaining 757 are distinct off-grid points.
The archive therefore combines a structured design with off-grid points,
rather than independent draws from a uniform distribution.

\begin{table}[htbp]
\centering
\caption{Biochemical inputs and stored ranges. Concentration units are reported
with the archive; the MAT file contains no unit metadata.}
\label{tab:ranges}
\begin{tabular}{lrrl}
\toprule
Input & Minimum & Maximum & Reported unit \\
\midrule
TF surface density, $\sigmatf$ & 100 & 10000 & Simulator-native; unverified \\
Fibrinogen, Fg & 4860 & 10310 & nM \\
Factor IX & 49.00 & 130.91 & nM \\
Factor VIII & 0.36 & 1.04 & nM \\
Factor V & 11.43 & 28.57 & nM \\
\bottomrule
\end{tabular}
\end{table}

\begin{figure}[htbp]
\centering
\includegraphics[width=.76\textwidth]{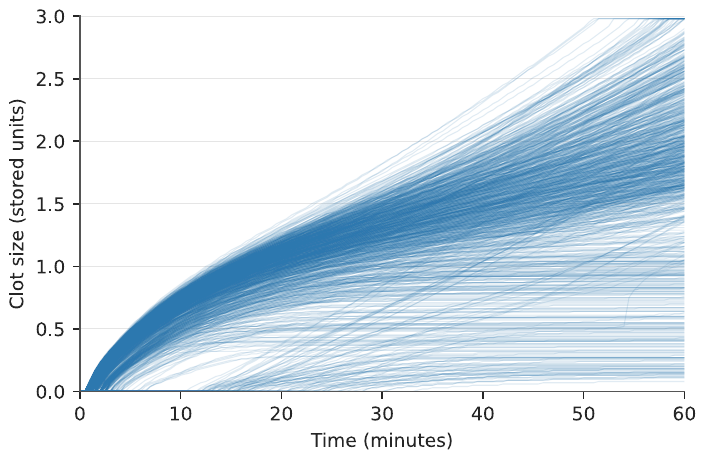}
\caption{All 1,000 archived clot-growth curves, without selection or rescaling.
Size is expressed in stored units.}
\label{fig:corpus}
\end{figure}

In the source model, $\sigmatf$ denotes tissue-factor surface density
on the activating cell surface, rather than a plasma volume concentration.
Clot size is the distance from that surface to the clot front, measured
perpendicular to it. The experimental front is defined by half-maximal
light scattering; the model supplies corresponding fibrin profiles.
The published growth curves report distance in millimetres and time in
minutes \cite[pp.~1490--1492]{panteleev2006spatial}. Stored sizes range
from 0 to 2.98, consistent with the model's 0--3\,mm spatial domain.
The published implementation used Watcom C/C++ 10.0 and an embedded
Runge--Kutta--Fehlberg method of order 2(3), adapted to the one-dimensional
reaction--diffusion equations. Surface reactions and plasma diffusion are
coupled through boundary flux conditions.

Some details needed to reproduce the clot simulations are unavailable:
the exact code snapshot, mesh, solver tolerances, and boundary settings for
these runs. The MAT file also does not explain how stored size corresponds
to the published clot-front definition. The numerical unit and any scaling
of stored tissue-factor surface density remain unverified. We therefore
evaluate forecasts in the stored units. The separate reaction benchmark
below is fully specified and has an independent numerical reference.

\subsection{Incomplete observations}
The forecasting task uses four early clot-size measurements to predict the
remainder of the curve. Let $M\subseteq\{1,\ldots,5\}$ index unknown biochemical inputs and
$K=M^c$ index known inputs. Four early size observations are taken at stored
times nearest 5, 12, 20 and 27 minutes. The target is the uncorrupted curve at
the $n_F=|F|=61$ stored times from 30 to 60 minutes. The same schedules are used for all
methods, including curves that have not begun to grow at the first visit.
The time windows define the benchmark rather than clinical decision deadlines.

In tests with exact biochemical measurements, $\theta_K$ is fixed and
$\theta_M$ is unavailable to the predictor. In tests with noisy measurements,
each measured factor has a specified independent Gaussian error, so its true
value remains uncertain during inference. In adaptive tests, all five inputs
start unknown. The method then chooses one of Fg, IX, VIII or V to measure;
tissue factor remains unmeasured. It chooses the assay using only the early
sizes and candidate predictions, before seeing the assay result.

\begin{table}[htbp]
\centering
\caption{Principal notation. Biochemical inputs are range-scaled when their
uncertainty or derivatives are compared.}
\label{mc:notation}
\begin{tabular}{ll}
\toprule
Symbol & Meaning \\
\midrule
$M,K$ & Missing and known input-index sets \\
$p,\theta,\xi$ & Model unknowns, biochemical inputs, trajectory-level latent vector \\
$\mathcal O,\mathcal Q$ & Early-observation and future-forecast maps \\
$\mathcal C_\varepsilon,D_\varepsilon$ & Compatible set and its forecast diameter \\
$B,w_b$ & Candidate count and normalized candidate weight \\
$s,T_w$ & Size-likelihood tolerance and weight temperature \\
$V_Q$ & Mean future variance under the candidate distribution \\
$r_j,A_j$ & Range-scaled assay noise SD and measurement-value score \\
$H_M,G_M,N$ & Early/future Jacobians and orthonormal unresolved basis \\
\bottomrule
\end{tabular}
\end{table}
\FloatBarrier

\section{Forecast ambiguity and numerical approximation}
\label{sec:forecast-framework}

\subsection{Information required for a forecast}
Unknown inputs need not prevent an accurate forecast. What matters is
whether inputs that fit the early observations also predict similar futures.
We first express this distinction without committing to a particular
surrogate or inference algorithm.

Let $p\in\mathcal P$ denote the unknowns of a specified model, including
trajectory-level latent variables when individual stochastic surrogate curves
are considered. Write $\mathcal O(p)$ for early predictions and
$\mathcal Q(p)$ for future predictions. For observations $y$ and tolerance
$\varepsilon$, define
\begin{equation}
 \mathcal C_\varepsilon(y)=\{p:\|\mathcal O(p)-y\|\le\varepsilon\},\qquad
 D_\varepsilon(y)=\sup_{p,q\in\mathcal C_\varepsilon(y)}
 \|\mathcal Q(p)-\mathcal Q(q)\|.
 \label{eq:forecast-diameter}
\end{equation}
Norms are fixed within each space; future errors use the root-mean-square
(RMS) norm over the forecast grid. The compatible set contains inputs whose
early predictions fit the observations within the tolerance. Its forecast
diameter is the largest difference between the futures predicted by those
inputs.

\begin{proposition}[Information limit]
\label{prop:forecast-diameter}
For a nonempty compatible set with finite diameter, every forecast $a$
satisfies
\begin{equation}
 \sup_{p\in\mathcal C_\varepsilon(y)}\|a-\mathcal Q(p)\|
 \ge \tfrac12D_\varepsilon(y).
\end{equation}
The forecast of any compatible parameter vector has worst-case error at
most $D_\varepsilon(y)$.
\end{proposition}
The triangle inequality gives the lower bound by comparing two compatible
explanations. The upper bound follows from the definition of diameter.
Thus ambiguous parameters can still support an unambiguous forecast if all
compatible inputs predict the same future. This statement concerns the
specified model and observation tolerance. It gives a worst-case information
limit rather than a probabilistic prediction interval.

\subsection{A bound for the weighted numerical forecast}
The information limit assumes that the model can be evaluated exactly.
A practical forecast also uses an approximate surrogate and a finite
candidate sample. The next bound separates these sources of error.

Let $\widehat{\mathcal O},\widehat{\mathcal Q}$ approximate reference maps
with uniform errors $\delta_O,\delta_Q$ on $\mathcal P$. Suppose the reference
maps are Lipschitz with constants $L_O,L_Q$ in a fixed parameter norm.

\begin{proposition}[Finite-candidate forecast error]
\label{prop:candidate-forecast}
Suppose $y=\mathcal O(p_*)+e$, $\|e\|\le\varepsilon$, and a candidate $p_b$
is within distance $h$ of $p_*$. Define
\[
 \tau=\varepsilon+\delta_O+L_Oh,\qquad
 A=\{k:\|\widehat{\mathcal O}(p_k)-y\|\le\tau\}.
\]
Let nonnegative weights sum to one and give positive mass to $A$.
Set $\alpha=\sum_{k\notin A}w_k$. If $D_A$ and $D_B$ are the approximate
forecast diameters in $A$ and in the full pool, respectively, then
\begin{equation}
 \left\|\sum_k w_k\widehat{\mathcal Q}(p_k)-\mathcal Q(p_*)\right\|
 \le (1-\alpha)D_A+\alpha D_B+\delta_Q+L_Qh.
 \label{eq:candidate-forecast-bound}
\end{equation}
\end{proposition}
Indeed $b\in A$, and
$\|\sum_kw_k\widehat{\mathcal Q}(p_k)-\widehat{\mathcal Q}(p_b)\|
\le(1-\alpha)D_A+\alpha D_B$.
Adding $\|\widehat{\mathcal Q}(p_b)-\mathcal Q(p_*)\|\le\delta_Q+L_Qh$
gives the result. The four terms have distinct meanings: disagreement among
accepted futures, weight assigned to candidates with poor early fits,
surrogate approximation error, and the effect of gaps in candidate coverage.
Using the bound requires constants that apply throughout the admissible
parameter set. Derivatives at sampled inputs and errors on validation cases
alone cannot establish those constants.

In the experiments, weighted forecast variance provides an easily computed
measure of candidate disagreement. With $Q_k=\widehat{\mathcal Q}(p_k)$ and
$\bar Q=\sum_kw_kQ_k$,
\begin{equation}
 V_Q=\sum_kw_k\|Q_k-\bar Q\|_{\rm RMS}^2
 =\frac12\sum_{k,l}w_kw_l\|Q_k-Q_l\|_{\rm RMS}^2.
 \label{eq:weighted-disagreement}
\end{equation}
This is average pairwise disagreement under the candidate weights. The
compatible-set diameter instead measures the largest disagreement. We test
whether the variance warns of forecast failure by comparing it with early-fit
residuals and errors on held-out curves.

The same distinction can be examined locally through derivatives. Write
$H_M$ and $G_M$ for the early and future Jacobians in range-scaled input
coordinates. Some small input changes leave the early predictions unchanged
to first order. Those changes also leave the future unchanged if and only if
$\ker H_M\subseteq\ker G_M$. With an orthonormal basis $N$ for the unresolved
input directions, $G_MN$ measures their effect on the future. The condition
$N^\top N=I$ makes its spectral norm independent of the chosen orthonormal
basis. Numerical rank thresholds estimate these local directions; they do
not describe compatible inputs far from the point being examined.

\subsection{Explicit growth-law examples}
Simple growth laws illustrate how a future trajectory can be determined
even when its underlying causes remain ambiguous. For
\begin{equation}
 \dot L=v+z,\quad \dot z=-\lambda z,\qquad
 L(t)=L_0+v\tau+z_0\frac{1-e^{-\lambda\tau}}{\lambda},
 \quad \tau=t-t_0,
 \label{mc:relaxation}
\end{equation}
take $L_0>0$, $v\ge0$, $z_0,\lambda>0$.
Four exact observations at increasing times determine $(v,z_0,\lambda)$
uniquely when an admissible curved solution exists. If $z_0=\kappa a_0$,
however, $(\kappa,a_0)$ and $(c\kappa,a_0/c)$ for $c>0$ produce the same whole curve.
The trajectory is therefore identifiable even though these two causes are
not. The exponential-fitting calculation is of Prony type
\cite{potts2010prony}. A unique solution for exact observations need not
be stable when observations are noisy. The growth law combines sustained
propagation with a temporary contribution that decays over time, motivated
by the distinction between initiation and propagation
\cite{ovanesov2005initiation,dashkevich2012thrombin}.

A second example shows how changing the model assumptions changes the
range of possible futures. A positive front with constant local supply and
a transport contribution that weakens as the front advances obeys
$\dot L=c+\beta/L$.
\begin{proposition}[Sharp envelopes and model dependence]
\label{mc:sharp-envelope}
For $0<L_0<L_1$ observed at $t_0<t_1$, every solution of
$\dot L=c+\beta/L$, $c,\beta\ge0$, through these observations satisfies
\begin{equation}
 \sqrt{L_1^2+h(L_1^2-L_0^2)}\le L(T)\le L_1+h(L_1-L_0),
 \qquad h=\frac{T-t_1}{t_1-t_0},\quad T\ge t_1.
 \label{mc:envelope}
\end{equation}
Both endpoints are attained, by pure transport and constant-speed growth.
If an unknown loss is allowed,
$\dot L=c+\beta/L-\gamma L$ with $c,\beta,\gamma\ge0$ and positive flow
through the observations, the sharp lower infimum at each $T>t_1$ becomes
$L_1$; the same linear upper bound is attained.
\end{proposition}
These envelopes bound future growth from two observations under the stated
growth laws. The surrogate experiments use four observations and a different
model family. The supplementary material gives the envelope proofs,
identification proofs, and numerical comparisons for the growth laws.

\subsection{Forecast value of a noisy measurement}
Inverse search can also help choose which missing factor to measure next.
We score an assay by the forecast uncertainty it could remove, accounting
for the assay's own measurement error.

Let $Q$ denote candidate future sizes and $X$ a range-scaled candidate factor.
Suppose a prospective assay gives $A=X+E$, with independent zero-mean error
of variance $r^2$. Set $c=\operatorname{Cov}(Q,X)$ and
$v_X=\operatorname{Var}(X)$.
\begin{proposition}[Affine measurement value]
\label{prop:affine-assay}
For finite second moments and $v_X+r^2>0$, the minimum mean squared risk over
affine forecasts $m(A)=\mathbb EQ+b(A-\mathbb EX)$ is
\begin{equation}
 \min_b\frac{\mathbb E\|Q-m(A)\|_2^2}{n_F}
 =V_Q-\frac{\|c\|_2^2}{n_F(v_X+r^2)},\qquad
 b=\frac{c}{v_X+r^2}.
 \label{mc:assay-risk}
\end{equation}
\end{proposition}
Expanding the quadratic and completing the square proves the identity.
The result requires independent assay error and finite second moments;
it does not require Gaussian distributions. The predicted risk reduction
defines our forecast-directed measurement score. With
$\eta_j=(\theta_j-a_j)/(b_j-a_j)$ denoting a range-scaled input, it is
\begin{equation}
 A_j^{\rm forecast}=\frac{n_F^{-1}\sum_{t\in F}
 \operatorname{Cov}_w[u(t),\eta_j]^2}{\operatorname{Var}_w(\eta_j)+r_j^2}.
 \label{mc:assay-score}
\end{equation}
Replacing future sizes by the range-scaled input vector gives a score aimed
at recovering the biochemical inputs. Both scores optimize an affine update
under the current candidate distribution. The actual procedure reruns
nonlinear inference after the assay, so an individual forecast can improve
or worsen. We assess this distinction by reporting gains and losses for
the same cases before and after measurement.

\section{Numerical forecasting procedures}
\label{mc:algorithms}

\subsection{Forward and direct predictors}
We use three forward surrogates to predict a full clot-size curve from
complete biochemical inputs. The stochastic neural ODE (SNODE) learns a
nonnegative growth rate. It also draws an eight-dimensional latent vector
from a conditional prior and keeps that vector fixed along the trajectory.
The archived implementation integrates the learned rate with fixed-step
Heun integration. The forward monotone MLP maps five inputs through four
192-unit layers with sigmoid linear unit (SiLU) activations to 121 positive
increments. The third surrogate combines
proper orthogonal decomposition with radial basis function interpolation
(POD--RBF). It interpolates coefficients of 12 trajectory modes learned from
the fitting curves, then projects the predicted curve onto nonnegative,
nondecreasing values.

These models test different representations of the same trajectories.
SNODE learns the growth dynamics, the forward MLP predicts increments
without integrating a differential equation, and POD--RBF uses nonlinear
interpolation in a low-dimensional trajectory space. Neural ODE and
latent-variable constructions are
established \cite{chen2018neuralode,rubanova2019latentode,kingma2022autoencodingvariationalbayes}.
The direct MLP provides a separate forecasting baseline. Its inputs are
early sizes, visit times, available biochemical values, and indicators of
which values are missing. It predicts future sizes in one evaluation,
without reconstructing missing inputs or searching a candidate pool.
Its architecture and selection procedure are specified with the numerical
comparison.

The supplementary comparisons also include a stochastic neural functional
differential equation (SNFDE), whose learned dynamics depend on past states.
Functional and neural delay equations provide its modelling framework
\cite{hale1993fde,zhu2021neural}. The Mori--Zwanzig representation explains
how eliminating unresolved states can introduce memory
\cite{chorin2000optimal,gouasmi2017memory}; the learned delay taps here are
not a memory kernel derived from the coagulation equations.
The latent-context control draws on conditional neural-process ideas
\cite{garnelo2018cnp}. Other controls use deep ensembles
\cite{lakshminarayanan2017deepensembles}, nearest-neighbour
retrieval \cite{altman1992kernelnearest}, principal-component representations
\cite{jolliffe2002pca} with ridge regression \cite{hoerl1970ridge}, and
Gompertz-curve retrieval \cite{gompertz1825mortality}.
These models represent trajectories and uncertainty in different ways.
Their earlier evaluation used a separate protocol, which is reported in
the supplementary material.

\subsection{Joint candidate integration}
Inverse search samples the missing inputs, evaluates their trajectories,
and weights them by agreement with the early observations. By default,
missing inputs have independent uniform priors over the fitting ranges,
forming the box $[a_M,b_M]$. A scrambled Sobol sequence, a randomized
low-discrepancy design, supplies $B=2^m$ candidates. Nested powers of two
preserve the sequence's balance; no points are skipped or thinned
\cite{owen2020dropping}. SNODE draws an additional latent vector for each
biochemical candidate and holds it fixed along that trajectory.

\begin{numericalalgorithm}{Weighted-candidate forecast}
\label{mc:algorithm-candidates}
\textbf{Inputs:} known values $\theta_K$; early observations $(t_j,y_j)$;
forward surrogate; prior on $\theta_M$; $B=2^m$, $s>0$, $T_w>0$; RNG seeds.
\begin{enumerate}
\setlength{\itemsep}{2pt}
\item Generate $B$ scrambled Sobol points in $[0,1]^{|M|}$ and transform
them to the missing-input prior. Keep $\theta_K$ fixed.
\item Evaluate each candidate curve $u_b$. For SNODE, draw its latent
vector from the conditional prior once per candidate.
\item Compute $\ell_b=-\sum_j[u_b(t_j)-y_j]^2/(2T_ws^2)$ and
$w_b=\exp[\ell_b-\operatorname{logsumexp}(\ell)]$.
\item Return weighted forecast/input means, variances and
$\mathrm{ESS}=1/\sum_bw_b^2$. Repeat with independent scrambles when
estimating numerical variation.
\end{enumerate}
\textbf{Output:} a finite working distribution and its forecast, conditional
on the supplied information and numerical budget.
\end{numericalalgorithm}

The effective sample size (ESS) summarizes weight concentration. A small
value means that a few candidates carry most of the weight.

All missing inputs vary together. A grid with $n_g$ values for each unknown
would instead require $n_g^{|M|}$ evaluations. The default size tolerance
$s=0.02$ in stored units accommodates surrogate mismatch even when the
observations are exact. It is a computational tolerance, not a measured
laboratory noise level. When size observations are noisy, their additional
variance is combined with this tolerance. The weight temperature is one
unless stated otherwise. When an alternative prior is evaluated on a uniform
candidate bank, its log density is added to $\ell_b$ before normalization.

\subsection{Conditioning on uncertain assays}
An uncertain assay narrows the plausible range of a factor without fixing
its true value. For an independent Gaussian assay of factor $j$, let $a_j^{\rm obs}$ be
its reported value and $\sigma_j$ its error standard deviation in raw units. Under a
uniform bounded prior, its assay-conditioned density is a normal
$N(a_j^{\rm obs},\sigma_j^2)$ truncated to $[a_j,b_j]$.
Writing $\Phi$ for the standard normal cumulative distribution function
(CDF), define
\[
 c_j=\Phi((a_j-a_j^{\rm obs})/\sigma_j),\qquad
 d_j=\Phi((b_j-a_j^{\rm obs})/\sigma_j).
\]
The transformed Sobol coordinate is
\begin{equation}
 \theta_j=a_j^{\rm obs}+\sigma_j
 \Phi^{-1}\{c_j+q_j(d_j-c_j)\}.
 \label{mc:assay-transform}
\end{equation}
We use stable tail formulas when direct CDF subtraction loses precision.
An exact assay fixes the factor value; an unmeasured factor keeps its uniform
prior.

\begin{numericalalgorithm}{Assay-conditioned proposal and forecast}
\label{mc:algorithm-assay}
\textbf{Inputs:} bounds; available exact/noisy assays; early sizes;
surrogate; numerical budget and RNG seeds.
\begin{enumerate}
\setlength{\itemsep}{2pt}
\item Generate scrambled Sobol coordinates for uncertain inputs.
\item Apply Eq.~\eqref{mc:assay-transform} to noisy measured coordinates,
uniform transforms to unmeasured coordinates, and fixed values to exact assays.
\item Evaluate forward trajectories, including prior latent draws for SNODE.
\item Apply the size-observation likelihood from Algorithm~\ref{mc:algorithm-candidates}
and return weighted moments.
\end{enumerate}
The proposal already contains the assay likelihood. Applying that likelihood
again would condition on the same measurement twice.
\end{numericalalgorithm}

This is standard inverse-CDF sampling from the assay-conditioned prior.
It concentrates candidates on biochemical values consistent with the assay.
The reaction benchmark checks both the transform and the subsequent
weighted inference against an independent numerical integral.

To choose an additional assay, we first use
Algorithm~\ref{mc:algorithm-candidates} to compute the covariances in
Eq.~\eqref{mc:assay-score}. We then measure the chosen factor and rerun
Algorithm~\ref{mc:algorithm-assay}. Random selection and always measuring IX
provide controls with the same initial and post-assay budgets. The policies
share the same potential assay errors and see an assay value only after
choosing its factor. Random-number seeds are independent of the policy,
so identical choices reuse the same post-assay candidate pool.

\section{Evaluation protocol}
\label{mc:evaluation}

\subsection{Fitting, selection, calibration and evaluation}
Each principal clot comparison uses three archived data splits (seeds
42--44). A split contains 800 fitting cases, 100 model-selection cases,
and 100 held-out cases. A fixed permutation, using seed plus 91,000,
divides the held-out group into 50 calibration cases and 50 evaluation
cases. Calibration sets interval widths; evaluation measures forecast
performance. These groups are disjoint within a split. Scaling and fitted
priors use only fitting data.

The saved SNODE selects its checkpoint using the original 100 selection
cases. The forward MLP, POD--RBF, and direct predictor instead select
settings on an 80/20 division of the 800 fitting cases, then refit on all
800. The original selection cases are used in the separate tolerance-selection
experiment, and are never used to calibrate intervals.

Across the splits, 150 evaluation appearances represent 143 distinct curves.
Independent scrambles repeat the candidate calculation on the same curves.
We average each metric over scrambles within a split, then give equal weight
to the three split summaries. Comparisons before and after an intervention
use the same cases. Although fitting, calibration, and evaluation have
separate roles, the archive was inspected during model development, so the
analyses remain exploratory. The supplementary material reports the earlier
ten-split, seven-model comparison and all 31 nonempty missing-input patterns,
with their separate calibration protocols.

\subsection{Errors and uncertainty}
We measure forecast error relative to the magnitude of the stored curves.
For a group of evaluation cases, pooled relative forecast error is
\[
 E_{\rm pool}=
 \left(\frac{\sum_i\sum_{t\in F}[\widehat u_i(t)-u_i(t)]^2}
 {\sum_i\sum_{t\in F}u_i(t)^2}\right)^{1/2}.
\]
Case-relative error applies the same formula to one curve. Both errors
measure agreement with simulation outputs rather than clinical risk.
Biochemical input errors are divided by their fitting-set ranges so that
factors with different numerical scales can be compared.

All principal interval comparisons use the same calibration rule. At each
future time, we divide each absolute calibration residual by the model's
base standard deviation (SD). The 46th smallest ratio among the 50
calibration cases sets the interval multiplier. For search models, the base
SD is the candidate SD, floored at $10^{-6}$ times the largest fitting size.
The direct model uses a base scale of one. Lower bounds are clipped at zero.
Evaluation outcomes do not affect the multipliers or the interval rule.
This shared calculation makes the reported intervals comparable while
preserving the original forecasts and data partitions. The finite-rank rule
is motivated by marginal residual calibration
\cite{angelopoulos2021conformal}. Here, the structured archive and
retrospective model development prevent a distribution-free coverage
guarantee.

We report pointwise coverage, mean width and the 90\% interval score
\cite{gneiting2007strictly}. For bounds $l,h$ and truth $u$, the score is
\[
 {\rm IS}_{0.1}(l,h;u)=(h-l)+20(l-u)_+ +20(u-h)_+.
\]
Smaller scores reward narrow intervals and penalize intervals that miss the
true value. Pointwise coverage is the fraction of individual future values
inside their intervals. Whole-curve coverage is the fraction of curves whose
entire future lies inside the band. These measure different events.
The numerical records retain whole-curve coverage and the results from the
original pooled calibration.

\subsection{Numerical and computational comparisons}
Candidate-budget checks increase the sample using nested prefixes of each
scrambled sequence. They use a 32,768-candidate calculation as a finite
reference, and variation across independent scrambles measures the remaining
sampling variability. The broad budget check uses 20 evaluation cases per
split (60 appearances, 59 distinct curves). The adaptive-assay and direct
comparisons use all 50. We report these groups separately.

Inference latency is the time to forecast one case after warm-up. It
includes candidate generation, surrogate evaluation, and weighted
aggregation. Training, model loading, and calibration fitting are excluded;
their costs are reported separately where recorded. GPU timing uses
synchronized calls. Each cost--accuracy plot compares models on the same
device and measures error on the cases that were timed. We do not compute
speed ratios between CPU and GPU measurements. The accompanying code package provides implementations, experiment
protocols, numerical tests, and a source-file hash manifest. Saved
predictions and other numerical outputs are not included in that package.

\paragraph{AI-assisted implementation.}
OpenAI ChatGPT and Codex assisted with the analysis, verification, and
plotting code. Executable calculations on the stored data or the specified
reaction benchmark produced the numerical results and figures. The accompanying code package provides implementations and automated
checks. Replaying the clot-growth analyses additionally requires the
simulation dataset and trained model checkpoints, which are not included
in the code package. The separate reaction benchmark can be run without
these files. The authors remain responsible
for scientific review.

\section{Numerical reliability and computational trade-offs}
\label{mc:results}

\subsection{Direct and inverse forecasts}

\label{sec:direct-baseline}

We first ask whether inferring the missing inputs improves forecast accuracy.
The direct baseline maps four observed sizes and their times, known
biochemical inputs, and indicators of missing inputs to 61 future sizes.
It predicts positive increments from the last observed size, without
candidate search or estimation of missing biochemical values.

The training procedure considers two SiLU architectures: three layers of
width 128 or four layers of width 192. Both use AdamW with learning rate
0.001, weight decay 0.00001, and batch size 128. Training runs for at most
1,200 epochs, with patience 160. At each epoch, every fitting case receives
a uniformly sampled nonempty missing-input pattern. Of the original 800
fitting cases, 640 fit the model and 160 select the architecture and epoch
count. Selection evaluates all 31 missing-input patterns every 20 epochs
and at epoch 1. The chosen model is then refitted on all 800 cases. Scaling
uses fitting cases only. The loss is normalized future-trajectory mean
squared error (MSE) plus half the terminal-time MSE. The direct model has
two architecture choices, while the existing forward MLP has one; the
methods also use different losses.

All three selected direct models have 126589 parameters. For seeds 42--44,
they are refitted for 720, 400, and 900 epochs, respectively. Selection and
refitting together take 0.12 to 0.24 minutes of CPU time per split. These
times exclude forward-surrogate training.

The direct baseline does not use the original 100 validation cases. Each
split retains separate groups of 50 calibration and 50 evaluation cases.
All methods use the same pointwise calibration rule: the 46th of 50 absolute
standardized residuals at each future time. The direct model uses unit
base scale, while search models use their candidate standard deviations.
The resulting bands calibrate marginal residuals; they do not make the
models' predictive distributions equivalent or guarantee coverage for
this structured archive.

\begin{table}[htbp]\centering\small

\caption{Matched direct and candidate-search forecasts. Entries average three split-specific results; candidate results with all five inputs unknown first average three scramble-specific metrics within each split, never their predictions. Each split has 50 evaluation cases (150 appearances, 143 distinct curves). Error is pooled relative $L_2$ in percent; coverage, width and interval score concern pointwise 90\% intervals. Width and score use stored size units.}

\label{tab:direct-baseline}

\begin{tabular}{llrrrrr}\toprule

Unknowns & Method & Candidates & Error (\%) & Cover & Width & Score \\

\midrule

TF & Direct MLP & -- & 1.50 & 0.903 & 0.071 & 0.095 \\

TF & Forward MLP & 2048 & 1.21 & 0.860 & 0.072 & 0.121 \\

TF & SNODE & 2048 & 1.37 & 0.889 & 0.081 & 0.121 \\

TF+Fg & Direct MLP & -- & 1.87 & 0.951 & 0.110 & 0.123 \\

TF+Fg & Forward MLP & 2048 & 1.52 & 0.870 & 0.071 & 0.120 \\

All five & Direct MLP & -- & 2.16 & 0.937 & 0.121 & 0.149 \\

All five & Forward MLP & 2048 & 3.66 & 0.886 & 0.194 & 0.311 \\

All five & Forward MLP & 8192 & 3.25 & 0.902 & 0.165 & 0.237 \\

All five & SNODE & 2048 & 3.56 & 0.882 & 0.196 & 0.298 \\

All five & SNODE & 8192 & 3.36 & 0.894 & 0.164 & 0.235 \\

\bottomrule\end{tabular}\end{table}

For the TF+Fg setting, a separate timing comparison evaluates the direct
MLP and frozen forward MLP on four CPU threads. Timing includes feature
preparation or candidate generation, prediction, aggregation, and interval
application for one case. After five warm-up calls, 25 timed calls cycle
through the first five evaluation IDs. Training and calibration fitting
are excluded. This comparison measures the two methods on the same CPU;
it does not provide a CPU--GPU ratio or a speed-up over the simulator.

\begin{figure}[htbp]\centering\includegraphics[width=\textwidth]{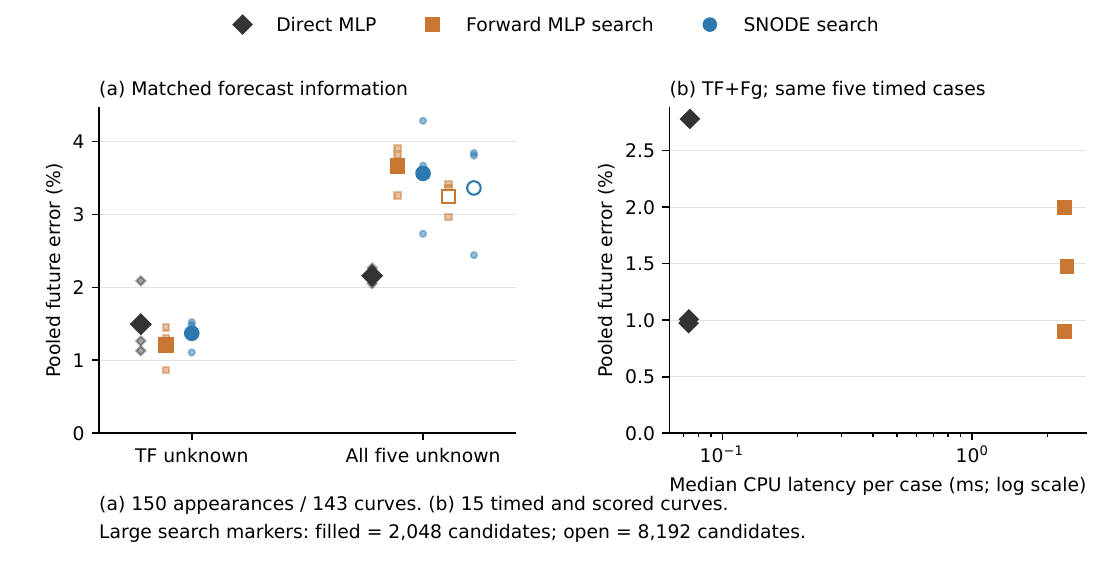}

\caption{Direct forecasting as a distinct computational strategy. (a) The original matched forecast comparison: three 50-case evaluation groups (150 appearances, 143 distinct curves); small points are split summaries and larger points their means. All-five search points average metrics over three scrambles. Large filled search markers use 2,048 candidates; large open markers use 8,192. (b) CPU-only TF+Fg cost/error comparison on exactly the five timed evaluation cases per split (15 distinct curves overall). Each point is one split: error pools its five cases, while median batch-one latency summarizes 25 calls cycling through those same five cases. Panel (b) therefore uses a smaller cohort than panel (a) and the table.}\label{fig:direct-baseline}\end{figure}

\begin{figure}[htbp]
\centering
\includegraphics[width=\textwidth]{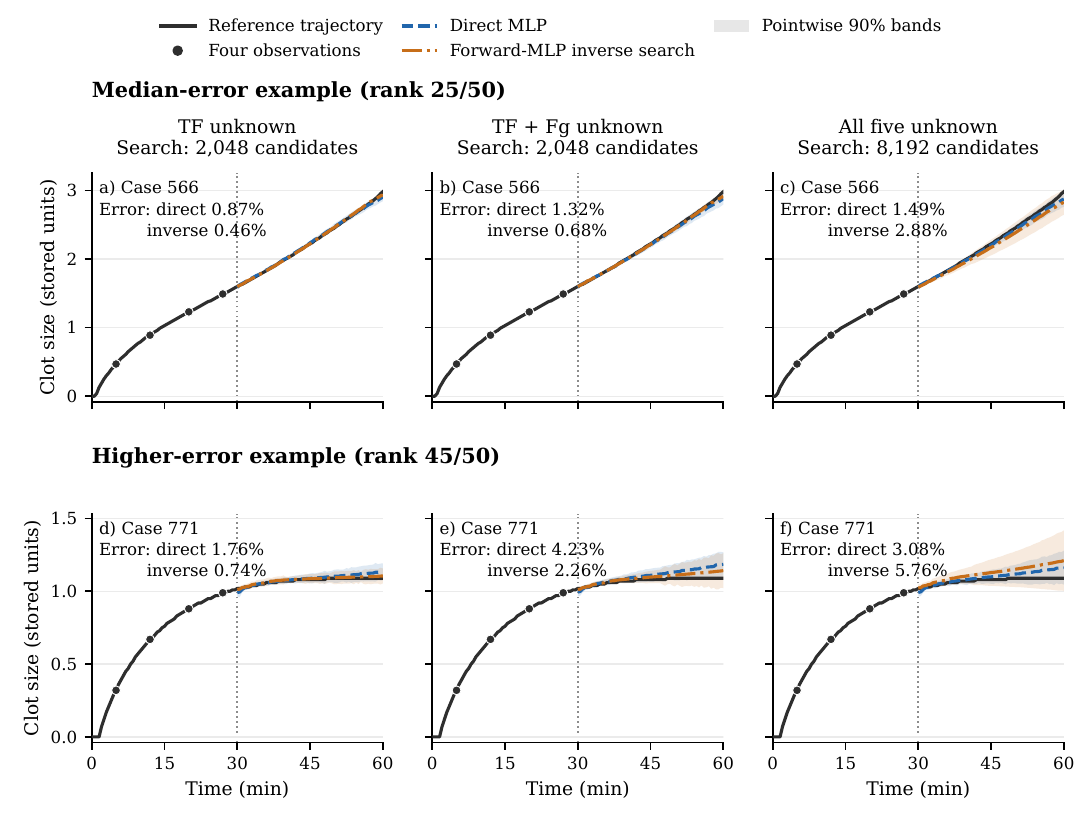}
\caption{Held-out forecast examples under three biochemical-information settings. Rows show the same two cases across settings, with shared vertical scales within each row and different scales between rows. Cases were selected post hoc at ranks 25 and 45 of the 50 evaluation cases in split seed 42, ordered by mean case-relative $L_2$ error across both methods and all three settings; case IDs are zero-based archive indices. Points mark the four observations and the dotted line marks the 30-minute forecast boundary. Shading gives pointwise 90\% residual-calibrated bands fitted on the separate 50-case calibration group. Error labels are case-relative $L_2$ percentages over 30--60 minutes. Search uses 2,048 candidates for TF and TF+Fg, and 8,192 for all five unknown inputs; the latter displays the first scramble (index 0), without averaging predictions.}
\label{fig:forecast-examples}
\end{figure}

\paragraph{Example trajectories.}
Figure~\ref{fig:forecast-examples} illustrates the forecasts for two held-out
curves. It compares the direct MLP with forward-MLP inverse search, the pair
tested in all three information settings. Both closely track the sustained
growth of case 566. Case 771 approaches a plateau, while both methods
continue to predict growth. The discrepancy and wider search band are
especially visible when all five inputs are unknown. The examples show
what the trajectory errors and interval widths look like for individual
curves. Table~\ref{tab:direct-baseline} reports the comparison over the
complete evaluation groups.

Table~\ref{tab:direct-baseline} and Figure~\ref{fig:direct-baseline}
show the main result: inverse search has lower average error when TF alone,
or TF and Fg, are unknown, while direct prediction has lower error when all
five inputs are unknown. The available biochemical information therefore
changes the relative accuracy of the two strategies. Since both the number
and identity of missing factors change, this result does not identify a
universal dimensional threshold.

The TF+Fg CPU comparison illustrates the cost of the accuracy gain. Median
time per forecast is approximately 0.074 ms for the direct model and
2.34--2.40 ms for MLP search on the timed subset. Search is more accurate
in that setting. It also returns plausible biochemical explanations and
supports measurement selection, which the direct baseline does not provide.
The value of these outputs, as well as forecast accuracy, determines whether
the extra cost is justified.

The remaining experiments examine the reliability of inverse search.
We first verify its numerical calculations, then test sensitivity to
candidate budget, solver, prior, and tolerance. Finally, we examine how
additional biochemical measurements affect the forecasts.

\subsection{Verification on a known reaction cascade}
\label{sec:mc-manufactured}
To verify the inference calculation independently of a learned surrogate,
we use a reaction system with known equations. Product concentration plays
the role of the observed trajectory, and a separate numerical integral
provides the reference forecast.

The irreversible mass-action cascade $A\to B\to C$ satisfies
$\dot A=-k_1A$, $\dot B=k_1A-k_2B$ and $\dot C=k_2B$, with
$(A,B,C)(0)=(m,0,0)$. It conserves $A+B+C=m$ and has nonnegative
solutions. The product concentration is
\begin{equation}
 C(t)=m\left[1-e^{-k_1t}-\frac{k_1}{k_2-k_1}
 (e^{-k_1t}-e^{-k_2t})\right].
 \label{eq:mc-cascade}
\end{equation}
All units are nondimensional. We infer all three inputs under independent
uniform priors: $k_1\in[0.1,0.5]$, $k_2\in[0.8,1.5]$ and $m\in[0.8,1.2]$.
Eight fixed input vectors provide four observations of $C$ at
$t=0.4,0.8,1.2,1.6$; forecasts are evaluated on 41 points from $t=2$ to 6.
Truth is generated independently by DOP853 with relative tolerance $10^{-12}$
and absolute tolerance $10^{-14}$, while candidates use Eq.~\eqref{eq:mc-cascade}.
The maximum solver--formula discrepancy was $3.35\times10^{-13}$, and
the maximum mass-conservation residual was $5.55\times10^{-16}$.

We test three observation conditions: exact product concentrations,
concentrations with independent Gaussian noise of standard deviation 0.02,
and those same noisy concentrations plus an independent mass assay with
Gaussian standard deviation 0.04. The observation likelihood uses scale
0.02 in every condition, including the clean case. Thus even exact
observations retain a positive scoring tolerance. No noisy value is clipped.
For the assay condition, inverse-CDF sampling from a truncated normal
exactly combines the uniform mass prior with the assay likelihood. Both
benchmarks use the same weighting and averaging routines; two zero-padded
coordinates adapt the reaction inputs to the five-input clot procedure.

Each case uses eight independently scrambled Sobol designs with nested
budgets of 512, 2,048, 8,192, and 32,768. These repeats measure numerical
variation on the same cases. The independent reference uses tensor
Gauss--Legendre quadrature to integrate the joint observation and assay
likelihood over the original uniform box. We check the reference using
orders 32, 48, and 64 per coordinate. The last two orders agree in every
mean and SD coordinate within $2.78\times10^{-14}$ after normalization,
below the prespecified $10^{-7}$ threshold. Forecast RMS quantities are
normalized by the fixed upper prior mass 1.2. Parameter differences are
normalized by each prior range, then combined using RMS.

At 32,768 candidates, mean forecast differences from quadrature are
$7.50\times10^{-6}$, $7.39\times10^{-6}$, $9.12\times10^{-6}$ for clean,
noisy, and assay-conditioned observations, respectively. The corresponding
parameter-mean differences are $4.09\times10^{-5}$, $4.00\times10^{-5}$, $4.62\times10^{-5}$.
These values measure numerical integration error. Error against the
generating inputs and trajectory is a separate quantity: even an accurately
integrated posterior mean may miss the truth when only four uncertain
observations are available. The benchmark verifies the numerical procedure;
it does not validate the coagulation model or clinical prediction.

\begin{figure}[tbp]
\centering
\includegraphics[width=\textwidth]{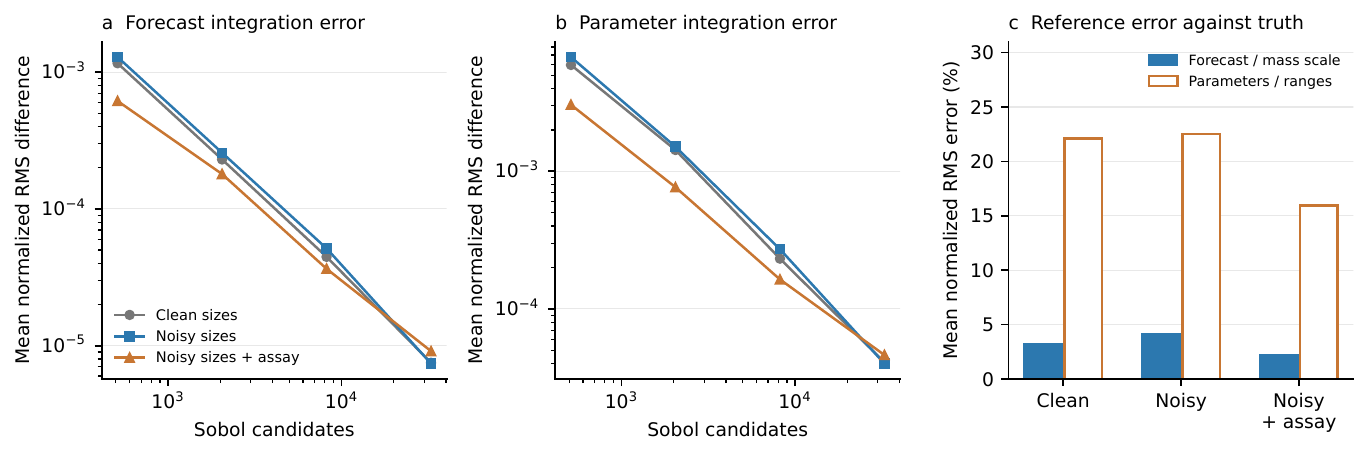}
\caption{Known-equation end-to-end verification. (a,b) Mean discrepancy from
the order-64 tensor-quadrature posterior mean over eight cases and eight
Sobol scrambles; reference accuracy is assessed by finite order refinement.
(c) Error of that reference posterior mean against the
generating truth, averaged over eight cases. Forecast and parameter errors
have different stated normalizations; the bars are not interchangeable
physical quantities. Clean observations retain a positive scoring tolerance.}
\label{fig:mc-manufactured}
\end{figure}

\paragraph{An exact instance of the finite-candidate bound.}
We also check the finite-candidate bound in a problem where all its constants
are known exactly. Let $L'=p$, $L(0)=0$, $p\in[0,1]$,
and $\widehat L(t;p)=(p+0.015)t$. With RMS output norms,
$L_O=\|t_O\|_{\mathrm{RMS}}$ and $L_Q=\|t_F\|_{\mathrm{RMS}}$ are exact
global Lipschitz constants, and $\delta_O=0.015L_O$,
$\delta_Q=0.015L_Q$ are uniform approximation errors. A midpoint grid of
$N$ candidates covers the entire input interval with radius $h=1/(2N)$.
At observation times $(0.2,0.4,0.6,0.8)$, alternating errors
$(0.01,-0.01,0.01,-0.01)$ have RMS magnitude $\varepsilon=0.01$.
Gaussian weights with scale 0.025 define the weighted forecast, evaluated on
21 future times from 1 to 2. For any accepted subset, its diameter is exactly
$(\max p_k-\min p_k)L_Q$, and the full-pool diameter is obtained similarly.
Substitution into Proposition~\ref{prop:candidate-forecast} verifies the
bound for $p_*=0.13,0.43,0.81$ and $N=8,16,32,64$: all 12 cases satisfy it.
The smallest nonnegative slack, the bound minus the actual error, was
$8.439\times10^{-2}$. Here the uniform constants and covering radius follow
directly from the equations and grid. This verifies the bound in the stated
problem; it does not establish the same assumptions for the clot benchmark.

\subsection{Candidate budgets and integration steps}
\label{mc:budget}
\begin{figure}[tbp]\centering
\includegraphics[width=\textwidth]{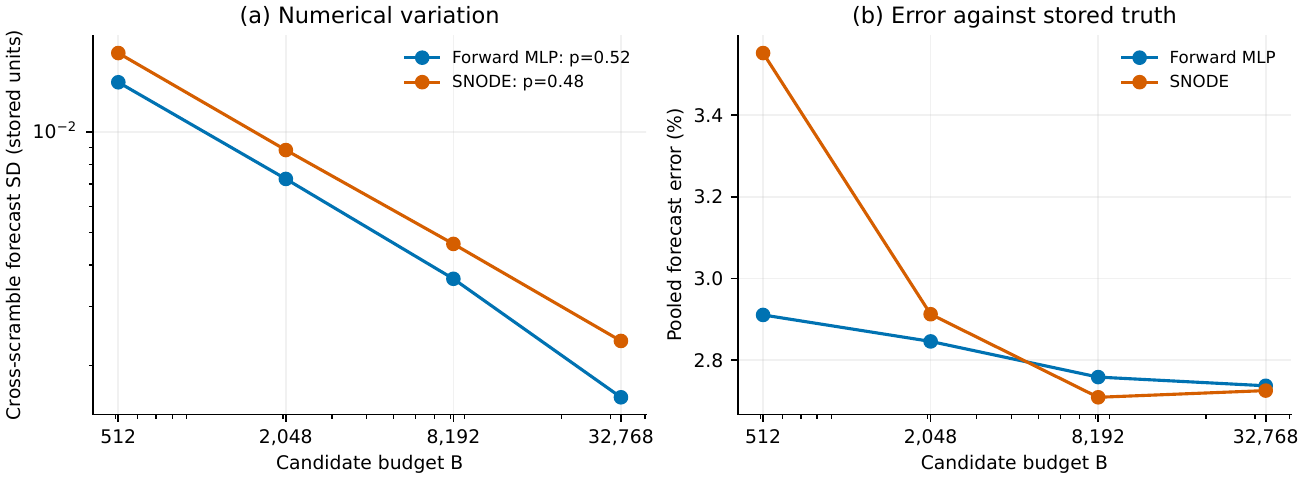}
\caption{Budget refinement on the same 60 evaluation appearances (59 distinct
curves). (a) Mean casewise RMS cross-scramble SD of forecast means, computed from three
scrambles for each case. Fitted exponents describe these four budgets only.
(b) Mean split/scramble pooled error against the stored trajectory.
SNODE variation includes both biochemical and latent sampling.}
\label{mc:budget-figure}
\end{figure}

This experiment tests whether a larger candidate sample improves forecast
stability and accuracy. We use nested budgets of 512, 2,048, 8,192, and 32,768 with all
five inputs unknown. Both surrogates use the same 20 evaluation cases per
split and three independent scrambles. Figure~\ref{mc:budget-figure} shows
that stability and accuracy behave differently. Increasing the budget
reduces average forecast SD across scrambles, but prediction error does
not decrease monotonically. At 32,768 candidates, mean pooled errors are
2.74\% for the MLP and 2.73\% for SNODE. At 2,048, they are 2.85\% and
2.91\%, respectively. Neither surrogate is consistently more accurate
across these budgets.

A log--log fit of mean forecast SD across scrambles against budget gives exponents
$p=0.520$ for the MLP and $0.476$ for SNODE in the descriptive relation
$\mathrm{SD}\propto B^{-p}$. Split-specific exponents span 0.491--0.536 and
0.431--0.541, respectively. The slopes are close to $B^{-1/2}$ over the
tested range. Four budgets and three scrambles provide limited evidence;
they do not establish an asymptotic rate or explain it through effective
dimension. The 2,048-candidate forecasts differ from their 32,768-candidate
references using the same scramble by mean case-relative RMS changes of
2.54\% (MLP) and 3.57\% (SNODE). These changes are appreciable relative
to the forecast errors. Even at the largest budget, SDs across scrambles
remain 0.00161 and 0.00237 stored units. Numerical stability and prediction
accuracy therefore require separate checks.

The comparison across all missing-input patterns also shows greater weight
concentration when more inputs are unknown. At the same 2,048-candidate
budget, SNODE's mean effective sample size drops from roughly 464 with
one missing input to 73 with all five missing. This is consistent with
more concentrated likelihood weights in the larger search space. It is an
observed pattern, not a general law of dimensional complexity. The
supplementary material reports every missing-input pattern.

SNODE also requires numerical integration of its learned dynamics. To
isolate this error, we fix its weights and use 20 inputs per split with
32 fixed latent draws per input. Heun integration with one, two, and four
substeps differs from fourth-order Runge--Kutta (RK4) integration with four
substeps by mean candidate RMS values
$7.00\times10^{-3}$, $2.01\times10^{-3}$ and $3.73\times10^{-4}$, normalized
by the fitting maximum size. This test concerns the learned field and keeps
candidate search fixed. It does not establish convergence of the original
mechanistic simulator. A solver change for a trained surrogate would also
require checking its forecast error.

\subsection{Working-prior sensitivity and failure diagnostics}

\label{mc:diagnostics}

Inverse forecasts depend on the prior as well as the observations. We test
that dependence with all five inputs unknown and 8,192 candidates. The
three priors are a uniform box, independent beta distributions fitted by
moments, and a fitted joint Gaussian restricted to the box. Prior fitting
uses only the 800 fitting inputs. The Gaussian covariance uses 10\%
diagonal shrinkage. We combine each prior with size tolerances
$s\in\{0.01,0.02,0.04\}$. All nine combinations use the same three
biochemical Sobol banks per split, with independently sampled latent
vectors added for SNODE.

\begin{table}[htbp]\centering\small

\caption{Working-prior sensitivity at fixed $s=0.02$. Error is pooled relative $L_2$ in percent. Coverage and interval score use the common pointwise calibration rule; scores use stored size units. Entries average three split summaries after averaging three scrambles within each split.}

\label{mc:prior-table}\begin{tabular}{llrrr}\toprule Model & Prior & Error & Coverage & Score \\\midrule

Forward MLP & Uniform & 3.31 & 0.904 & 0.236 \\

Forward MLP & Marginal beta & 3.18 & 0.908 & 0.236 \\

Forward MLP & Joint Gaussian & 3.55 & 0.900 & 0.246 \\

SNODE & Uniform & 3.41 & 0.898 & 0.236 \\

SNODE & Marginal beta & 3.29 & 0.901 & 0.234 \\

SNODE & Joint Gaussian & 3.64 & 0.906 & 0.245 \\

POD--RBF & Uniform & 4.28 & 0.878 & 0.290 \\

POD--RBF & Marginal beta & 4.18 & 0.891 & 0.293 \\

POD--RBF & Joint Gaussian & 4.50 & 0.867 & 0.298 \\

\bottomrule\end{tabular}\end{table}

The forward MLP has lower aggregate error than SNODE for every prior at
$s=0.01$ and $0.02$. At $s=0.04$, SNODE has slightly lower error, by
0.02--0.04 percentage points. Both have lower error than POD--RBF in all
nine comparisons. At $s=0.02$, the beta priors improve mean error for all
three models, while the joint Gaussian worsens it. These priors summarize
the simulation design, not a patient population.

Changing the prior also changes the recommended assay. Relative to the
uniform prior, the two alternatives change the forecast-directed factor
choice on 17.3--25.3\% of evaluation appearances for the forward MLP,
18.4--27.1\% for SNODE, and 18.7--28.4\% for POD--RBF. Thus measurement
choices can remain sensitive to the prior even when the model ordering
stays the same.

In a separate tolerance-selection experiment, we minimize mean forecast
MSE on the original 100 validation cases, averaging across scrambles.
Calibration and evaluation outcomes are not used for this choice. The
numerical records include the full tolerance grid and the results for
the selected tolerance.

\begin{figure}[tbp]\centering\includegraphics[width=\textwidth]{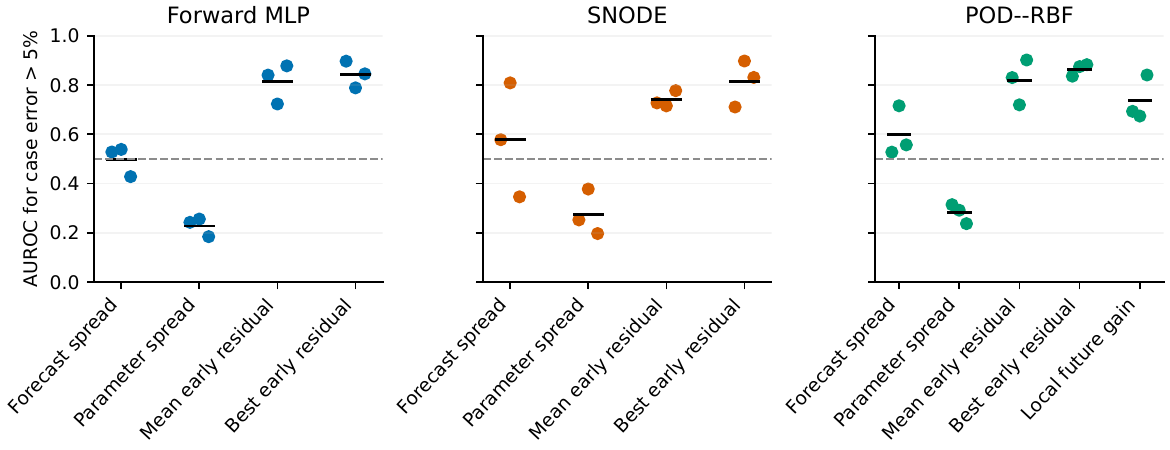}

\caption{Forecast-failure diagnostics with all five factors unknown, a uniform prior and $s=0.02$. A failure is mean case-relative $L_2>5\%$, averaging errors over three scrambles, not averaging forecasts. Scores also average those three scrambles per case. Each dot is one 50-case split; black marks are descriptive means. The dashed line is chance discrimination. Local future gain is evaluated only for deterministic POD--RBF. Repeated appearances are not independent cases.}\label{mc:diagnostic-figure}\end{figure}

We next ask which quantities warn that a forecast is likely to fail
(Figure~\ref{mc:diagnostic-figure}). Failure means case-relative $L_2$
error above 5\%, averaged over three scrambles. Discrimination is measured
by the area under the receiver operating characteristic curve (AUROC);
0.5 indicates chance discrimination.

The early residual of the best-fitting candidate is a more consistent
warning than raw forecast spread. Its AUROC ranges are 0.79--0.90 for the
MLP, 0.71--0.90 for SNODE, and 0.84--0.88 for POD--RBF. Forecast spread
gives 0.43--0.54, 0.35--0.81, and 0.53--0.72, respectively. Parameter
spread has AUROC below 0.5 in every split. Candidate disagreement therefore
cannot be interpreted as a calibrated probability of failure.

For POD--RBF, the local gain in unresolved input directions has AUROC
0.67--0.84 at the fixed central finite-difference and rank thresholds
$10^{-3}$. It provides useful information, but is not consistently better
than an early-fit residual. Local derivatives and weighted candidate spread
measure different aspects of ambiguity. Neither accounts for the surrogate
error in Proposition~\ref{prop:candidate-forecast}.

\subsection{Conditioning on noisy biochemical measurements}
We next test whether forecasts improve when uncertain biochemical
measurements are treated as uncertain. TF remains unmeasured, while
independent Gaussian errors perturb Fg, IX, VIII, and V. Their SDs are
2\%, 5\%, or 10\% of the fitting ranges. Methods and noise levels share
the same potential errors, and the four size observations remain clean.
Table~\ref{tab:noisy-factors} compares fixing the reported factor values
with integrating over their assay-conditioned distributions. At 10\%
noise, accounting for assay uncertainty reduces pooled error from 4.96\%
to 2.84\% for SNODE and from 5.05\% to 2.60\% for the MLP. Both forward
surrogates therefore benefit.
\begin{table}[tbp]\centering\small
\caption{Uncertain biochemical measurements with tissue factor unmeasured, using 2,048 candidates. Assay SD is a fraction of each training factor range, not a coefficient of variation. Values average three split summaries with 50 calibration and 50 evaluation cases per split. Forecast error and TF range-normalized MAE are percentages. Coverage, width and interval score use the common pointwise 90\% band rule; widths and scores use stored size units. Exact uses true factors; fixed treats noisy measurements as exact; aware integrates assay uncertainty.}
\label{tab:noisy-factors}\begin{tabular}{llrrrrr}\toprule Model / noise & Rule & Error & TF MAE & Cover & Width & Score \\\midrule
SNODE / 0\% & Exact & 1.37 & 2.16 & 0.889 & 0.081 & 0.121 \\
SNODE / 2\% & Fixed & 1.68 & 2.60 & 0.895 & 0.108 & 0.143 \\
SNODE / 2\% & Aware & 1.63 & 2.58 & 0.907 & 0.089 & 0.119 \\
SNODE / 5\% & Fixed & 2.75 & 3.92 & 0.889 & 0.174 & 0.229 \\
SNODE / 5\% & Aware & 2.30 & 3.59 & 0.891 & 0.111 & 0.146 \\
SNODE / 10\% & Fixed & 4.96 & 6.59 & 0.893 & 0.316 & 0.407 \\
SNODE / 10\% & Aware & 2.84 & 4.75 & 0.914 & 0.135 & 0.191 \\
\midrule
MLP / 0\% & Exact & 1.21 & 1.60 & 0.860 & 0.072 & 0.121 \\
MLP / 2\% & Fixed & 1.56 & 2.08 & 0.864 & 0.105 & 0.156 \\
MLP / 2\% & Aware & 1.49 & 2.11 & 0.869 & 0.082 & 0.118 \\
MLP / 5\% & Fixed & 2.72 & 3.66 & 0.881 & 0.200 & 0.268 \\
MLP / 5\% & Aware & 2.18 & 3.32 & 0.859 & 0.103 & 0.150 \\
MLP / 10\% & Fixed & 5.05 & 6.53 & 0.898 & 0.382 & 0.488 \\
MLP / 10\% & Aware & 2.60 & 4.58 & 0.867 & 0.121 & 0.182 \\
\bottomrule\end{tabular}\end{table}

Lower error does not by itself imply better uncertainty coverage. The MLP
still falls below nominal coverage, so its narrower intervals must be
considered together with coverage and interval score. When all five true
inputs are supplied, forward-only pooled errors on these same cases are
1.34\% for SNODE and 1.17\% for the MLP. These values measure surrogate
approximation error with complete inputs. Subtracting them from inverse
forecast errors would not isolate a separate additive error contribution.

\subsection{Noisy adaptive measurements}
We now test whether the candidate distribution can guide the choice of one
additional assay. All five inputs start unknown. Each policy chooses one of
Fg, IX, VIII, or V using the same early observations, before the assay result
is revealed. The forecast-directed policy aims to reduce forecast uncertainty;
the parameter-directed policy aims to reduce uncertainty in the biochemical
inputs. Both use the noise-aware affine-risk scores. Random selection and
always measuring IX provide controls.

Policies, models, and budgets share the same simulated Gaussian assay errors.
An exact assay fixes the factor value. A noisy assay updates its prior while
retaining uncertainty about the true value. All policies use the same nominal
candidate budget before and after measurement. Policies choosing the same
factor reuse the same posterior calculation, avoiding duplicate work.

Each frozen split retains 50 calibration cases and 50 separate evaluation
cases, neither used for fitting or checkpoint selection. Pointwise 90\%
intervals use the common rule: the 46th ordered standardized absolute
calibration residual, an SD floor of $10^{-6}$ times the training maximum,
and a lower bound clipped at zero. We average metrics over three independent
candidate/latent sampling repetitions within each split. The 150 evaluation
appearances represent 143 distinct stored curves.
\begin{table}[htbp]\centering
\caption{Adaptive-assay forecast errors (percent pooled relative $L_2$), averaged over three split-specific summaries after averaging three search repetitions within each split. Noise is a fraction of the factor training range. No-assay, forecast-directed, parameter-directed, random and fixed-IX rules are retained. The 5\% noise level was tested only at 2,048 candidates.}
\label{tab:matcom-adaptive}
\begin{tabular}{llrrrrr}\toprule
Model & $B$ / noise & None & Forecast & Parameter & Random & IX \\ \midrule
SNODE & 2,048 / 0\% & 3.56 & 3.45 & 3.16 & 3.68 & 3.34 \\
SNODE & 2,048 / 5\% & 3.56 & 3.53 & 3.49 & 3.63 & 3.57 \\
SNODE & 2,048 / 10\% & 3.56 & 3.53 & 3.45 & 3.74 & 3.65 \\
SNODE & 8,192 / 0\% & 3.36 & 3.12 & 3.01 & 3.27 & 3.26 \\
SNODE & 8,192 / 10\% & 3.36 & 3.16 & 3.23 & 3.53 & 3.31 \\
\midrule
MLP & 2,048 / 0\% & 3.66 & 3.25 & 3.33 & 3.52 & 3.48 \\
MLP & 2,048 / 5\% & 3.66 & 3.07 & 3.15 & 3.36 & 3.41 \\
MLP & 2,048 / 10\% & 3.66 & 3.09 & 3.15 & 3.38 & 3.47 \\
MLP & 8,192 / 0\% & 3.25 & 2.87 & 3.21 & 3.21 & 3.29 \\
MLP & 8,192 / 10\% & 3.25 & 2.84 & 2.96 & 3.07 & 3.29 \\
\bottomrule\end{tabular}\end{table}
At 10\% assay noise and $B=8192$, the forecast-directed policy has lower
mean error than random selection for both models. SNODE's paired
forecast-minus-random difference is -0.366 percentage points, with sample
SD 0.304 across three split summaries. For the MLP, the difference is
-0.237 percentage points with SD 0.193. These descriptive summaries are
not confidence intervals.

An assay does not improve every forecast. Compared with no assay, the
forecast-directed rule improves 55.1\% of SNODE predictions and worsens
44.9\%. It improves 58.0\% of MLP predictions and worsens 42.0\%. These
fractions average nine search runs on the same split-specific cases.
SNODE's pointwise coverage is 87.9\%, with mean interval width 0.144 and
score 0.226. The MLP's coverage is 88.2\%, with width 0.143 and score
0.215. Widths and scores use stored size units.

The choice of assay also depends on the candidate budget. At 10\% noise,
the two budgets choose the same factor in 62.0\% of SNODE and 67.6\% of
MLP evaluation searches. Ranking the four assay policies by mean forecast
error changes with the budget in 3 of the four model--noise comparisons
(0\% and 10\% noise). The two-model ranking changes in 2 of ten
policy--noise comparisons, including the no-assay reference. These are
observed numerical rankings; they do not establish statistical significance
or convergence.
\begin{figure}[htbp]\centering\includegraphics[width=\textwidth]{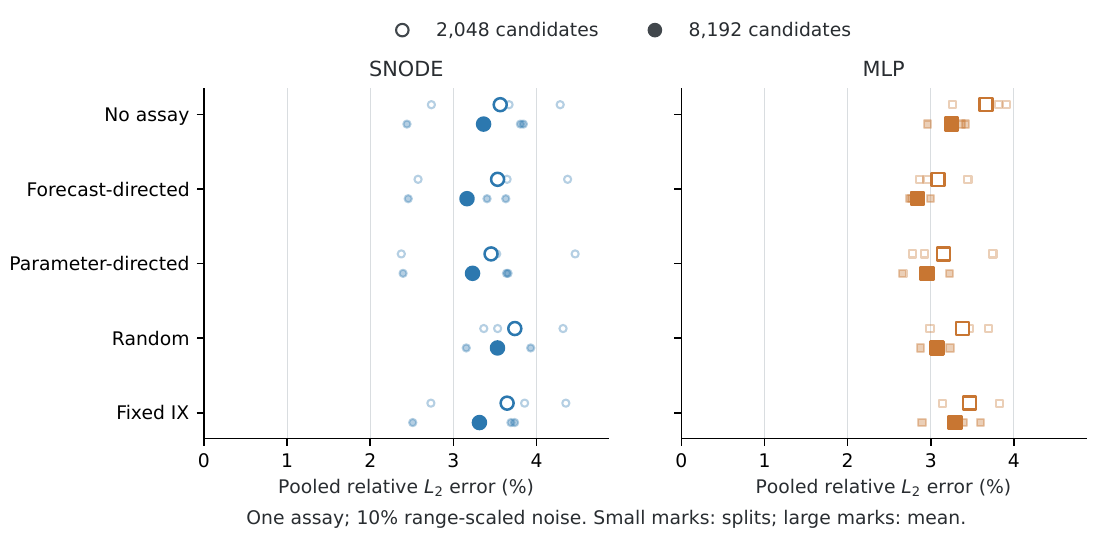}
\caption{One additional noisy assay at 10\% range-scaled noise. Open markers use 2,048 candidates and filled markers use 8,192, both before and after measurement. Small marks are three split-specific summaries; larger marks show their mean. Each split summary averages three searches, not their predictions. The no-assay reference uses only the initial pool. Model panels use the same 150 evaluation appearances.}\label{fig:matcom-adaptive}\end{figure}
\subsection{Measured cost on a fixed subset}
To measure the cost--accuracy trade-off, we forecast one case at a time on
one NVIDIA RTX PRO 6000 Blackwell Workstation Edition GPU. Both frozen
models and every budget use the first three predetermined evaluation cases
per split. Five search repetitions per case time candidate generation,
forward prediction, and weighting. Errors are measured on those same nine
distinct cases. All inputs are unknown and no additional assay is acquired
in this timing test.

Across the 45 calls at each budget, SNODE's largest allocated GPU-tensor
peaks are 14.9, 18.4, 23.1, 23.1 MiB at $B=512,2048,8192,32768$,
respectively. The MLP's peaks are 10.4, 12.7, 21.8, 58.3 MiB at
$B=512,2048,8192,32768$, respectively. These measurements exclude CPU
candidate arrays, reserved device memory, and total system memory.

The recorded forward-MLP selection, refit, and checkpoint-save times span
3.83--11.17 seconds per split. SNODE's archived logs record 14.05--21.91
minutes through the final training/validation epoch. These historical
timings used different training procedures and do not constitute a matched
training-cost comparison. SNODE checkpoint loading takes 0.006--0.121
seconds. Neither model is retrained for this benchmark. Calibration fitting
was not separately timed and is excluded from inference latency.
\begin{figure}[htbp]\centering\includegraphics[width=.88\textwidth]{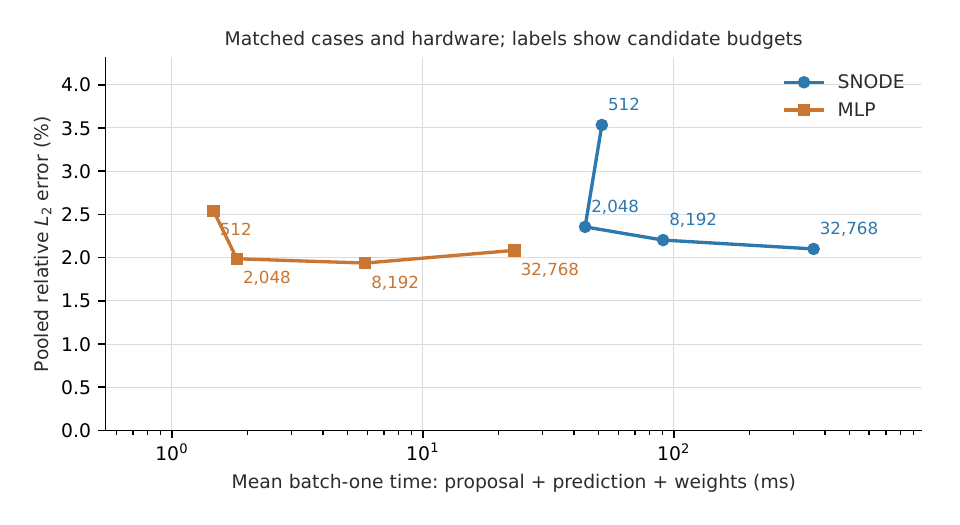}
\caption{Measured cost and forecast error on the same nine predetermined cases (three per split), using both frozen models on the same GPU. Each labelled point is a candidate budget. Horizontal position is mean synchronized batch-one time including proposals, prediction and weights over 45 calls; vertical position averages per-split/per-repetition pooled errors over the three cases. Training, calibration and laboratory time are excluded.}\label{fig:matcom-assay-cost}\end{figure}

\section{Discussion}

The available biochemical information changes which forecasting strategy
is more accurate. Inverse search has lower error when TF alone, or TF and
Fg, are unknown. Direct prediction has lower error when all five inputs
are unknown
(Table~\ref{tab:direct-baseline}). This reversal changes whether the
accuracy gain from inverse search justifies its cost. The TF+Fg timing
comparison illustrates the trade-off on a smaller group evaluated on the
same hardware. The conclusion applies to the tested missing-factor
patterns and training procedures. It does not identify a universal
threshold for the number of unknown inputs.

The biological motivation is that a short clot-size record may not
distinguish all processes that determine later growth. The source
reaction--diffusion study distinguishes initiation, propagation and
localization \cite{panteleev2006spatial}. This provides a reason to examine
both early observations and biochemical information, but it does not
establish a biological cause for the observed forecasting reversal.
The forecasts predict scalar simulation outputs. Their accuracy alone
cannot establish pathway activity or validate a candidate biochemical
explanation.

One possible explanation is computational. Known factor values restrict
the candidates that inverse search must consider. When all inputs are
unknown, the direct model can use relationships between early and future
growth learned from the archive, without candidate-sampling error. This
explanation has not been isolated experimentally. Training objectives and
selection opportunities also differ, the direct model has its own
approximation error, and larger search budgets do not consistently reduce
forecast error. The reversal therefore needs to be interpreted together
with the budget, prior, and training conditions.

Inverse search has uses beyond predicting future size: it returns candidate
biochemical explanations and can guide assay selection. The noisy-assay
experiments show why these outputs matter. Additional factor information
can improve a forecast, but treating a noisy assay as exact can increase
error. Choosing an assay to recover parameters also need not give the best
forecast. The appropriate strategy therefore depends on the required
outputs, available measurements, and computational budget.

The mathematical framework separates three checks on reliability. The
reaction example verifies candidate integration against an independent
reference. Budget and solver comparisons measure numerical sensitivity
of the clot surrogates. Held-out diagnostics then test whether candidate
disagreement and local ambiguity warn of forecast errors. In those tests,
raw candidate spread is less reliable than an early-fit residual. The local
gain in unresolved input directions is useful, but is not consistently
better than that residual. Candidates can agree because they share the
same model error. Similarly, forecast error that grows with time does not
by itself identify a particular dynamical instability
\cite{strogatz2015nonlinear}.
Related studies of neural-operator failure modes and semigroup consistency
motivate diagnostic checks beyond aggregate prediction error
\cite{shikhman2026diagnosing,shikhman2026semigroup}.

The growth-law examples make the distinction explicit. Several underlying
causes can produce the same identifiable future. Allowing an additional
loss mechanism can instead enlarge the range of possible futures, even
with the same early observations. The analytical envelopes apply only
when those growth laws are imposed; they do not bound the neural candidate
forecasts. These examples explain why the assumed model family matters
and why low candidate variance does not resolve uncertainty about that
model family.

Biomedical learning also addresses tumour-growth forecasting
\cite{chen2023tgm,chen2025deep}, thrombotic-risk assessment
\cite{gao2023deep} and blood-glucose modelling \cite{deng2026recent}.
Blood-cell classification \cite{deng2023deep} and retinal-image segmentation
\cite{zhang2022aoslo} concern different prediction targets and measurements.
These studies provide methodological context, but do not supply validation
data for the clot-growth forecasts studied here.

\subsection{Scope and limitations}
All clot comparisons use one simulation archive generated with code from
an identified published model. The exact code snapshot and run configuration
remain unavailable. The structured parameter design, absence of repeated
physical measurements, and prior inspection during development limit the
interpretation of the results. The separate reaction cascade verifies
numerical inference, rather than coagulation biology.

The monotonicity constraint fits these stored curves but cannot represent
sustained clot regression or lysis. Fitting curves already span 0--60
minutes, so predicting a new curve over that window is not extrapolation
beyond the training time range. In the supplementary material, an
exploratory low-TF holdout gives poor forecasts for both refitted
deterministic controls.

The priors and independent Gaussian assay errors are computational
assumptions. Finite budgets, three scrambles, and different training
opportunities limit conclusions about model rankings and assay policies.
Sampled POD--RBF derivatives and observed errors also do not establish
uniform bounds for the reference simulator. Timing that simulator,
testing new parameter regimes, and evaluating experimental clot growth
would require resources beyond the archive. Patient-specific use would
also require appropriate measurement-error models and independent
clinical validation.

\section{Conclusion}
The relative accuracy of direct and inverse clot-growth forecasting depends
on which biochemical inputs are available. In the tested archive, inverse
search is more accurate when TF alone, or TF and Fg, are unknown; direct
prediction is more accurate when all five inputs are unknown. This observed
reversal concerns the tested information patterns and training procedures.

Inverse search remains useful when its accuracy gain, candidate biochemical
explanations, or support for assay selection justifies its cost. Its
reliability requires separate checks on observation ambiguity, surrogate
error, and numerical integration. The reaction benchmark verifies the
integration procedure, while the clot experiments show sensitivity to
budget, tolerance, and prior. Together, the comparisons provide a basis
for choosing a forecasting strategy according to the available information
and the outputs required.

\section*{Declarations}
\paragraph{Data and code availability.}
The accompanying code package contains analysis code,
figure-generation scripts, numerical verification tests, and the
reaction-cascade generator and reference quadrature.
The clot-growth trajectories analyzed in this study were generated
using code associated with the spatial coagulation model of
Panteleev et al.\ \cite{panteleev2006spatial}.
The clot-growth dataset and trained model checkpoints are not
included in the code package.

\paragraph{Funding.}
This work was supported by National Institutes of Health grants
R21HL168507 and R01GM163243, NSF SCH award 2406212, and
Dell Technologies. Dell Technologies provided computational resources
that supported the original study, including a Dell Pro Max T2
workstation and a Dell Pro Max 16 Plus laptop.
The funders had no role in study design, data collection, analysis or
interpretation, manuscript writing, or the decision to submit this
article for publication.
\paragraph{Author contributions.}
L.J.S.: methodology, software, validation, formal analysis, visualization,
writing--original draft, and writing--review and editing.
L.D.: formal analysis.
Y.Q.: conceptualization, methodology, investigation, and writing--review and editing.
H.L.: conceptualization, resources, supervision, project administration,
and writing--review and editing.
\paragraph{Competing interests.}
The authors declare no competing interests.
\paragraph{Ethics.}
This study uses synthetic trajectories and computational experiments.
It recruits no participants, uses no human specimens and analyses no
identifiable patient records. Informed consent does not apply.
No institutional ethics approval or exemption determination is claimed.

\paragraph{Declaration of generative AI and AI-assisted technologies in the manuscript preparation process.}
OpenAI ChatGPT and Codex assisted with coding. Codex also assisted
with revising the data and code availability statement and preparing
submission declarations. The authors reviewed the revised content and
take full responsibility for the code, results, and final manuscript.

\bibliographystyle{plainnat}
\bibliography{references_amm,references_matcom_additions}
\end{document}


\maketitle
\vspace{-2em}
\section{Biological and modelling context}
Platelet activation, thrombin regulation and fibrin formation are coupled
\cite{versteeg2013new,palta2014overview,furie2008mechanisms,butenas2002blood,tanaka2009blood}.
Experiments distinguish clot initiation from propagation and demonstrate
responses to combined stimuli
\cite{cawthern1998blood,ovanesov2002hemophilia,ovanesov2005initiation,chatterjee2010pairwise}.
Coagulation models span reaction thresholds, surface binding and thrombin
production
\cite{leiderman2014overview,anand2022computational,beltrami1995mathematical,kuharsky2001surface,bungay2003mathematical,hockin2002model}
and spatial transport, flow and platelet deposition
\cite{zarnitsina1996mathematical,fogelson2005coagulation,leiderman2011grow,leiderman2013influence}.
Traveling waves and models of initiation and growth under flow add
spatiotemporal structure
\cite{dashkevich2012thrombin,bouchnita2017mathematical,bouchnita2019multiscale,bouchnita2020mathematical,bouchnita2023thrombin}.
A single size curve cannot be assumed to identify all these processes.

Multifidelity and surrogate modelling can reduce repeated forward-simulation
cost \cite{guerrero2023multifidelity,gherman2023bridging}. Relevant approaches
include physics-informed coagulation models \cite{qian2024coagulonet},
neural and latent ODEs \cite{chen2018neuralode,rubanova2019latentode},
and surrogates for circulation, cell-based simulation, aortic flow and sparse
biological observations
\cite{yuhn2022uq,comlekoglu2025surrogate,daneker2024transfer,lagergren2020binn}.
Related learning applications address tumour growth
\cite{chen2023tgm,chen2025deep}, thrombotic risk \cite{gao2023deep},
glucose \cite{deng2026recent}, blood cells \cite{deng2023deep} and retinal
images \cite{zhang2022aoslo}; their different measurements do not validate
simulated clot-size prediction.

A biomarker review documents gaps between routine measurements and model
inputs \cite{albannoud2024biomarkers}. Neural networks have been coupled to
coagulation equations for kinetic-parameter estimation
\cite{albannoud2025kinetics} and to a hybrid recurrent-venous-thromboembolism
framework \cite{albannoud2026surrogates}. Arbitrary-subset conditioning is
also established \cite{ivanov2019arbitrary}. These provide context for
forecasting with incomplete inputs, not a basis for architectural novelty.
Error growth alone does not identify a dynamical instability
\cite{strogatz2015nonlinear}; learned-model failure and temporal-composition
studies motivate explicit checks
\cite{shikhman2026diagnosing,shikhman2026semigroup}.

\section{Forecast ambiguity and weighted-candidate proofs}
\label{S-sec:forecast-framework}
Let $p\in\mathcal P$ denote the unknowns of a specified model, with known
inputs fixed. For stochastic latent surrogates, $p$ also includes the
trajectory-level latent vector. Write $\mathcal O(p)$ and $\mathcal Q(p)$
for observation and future-output vectors. Norms are fixed within each
space; the future norm is RMS over the time-index set $F$, with $n_F=|F|$. Define
\begin{equation}
 \mathcal C_\varepsilon(y)=\{p:\|\mathcal O(p)-y\|\le\varepsilon\},\qquad
 D_\varepsilon(y)=\sup_{p,q\in\mathcal C_\varepsilon(y)}
 \|\mathcal Q(p)-\mathcal Q(q)\|.
 \label{S-eq:forecast-diameter}
\end{equation}
A prior can change the weight given to compatible explanations without
making them distinguishable from observations \cite{raue2009identifiability,stuart2010inverse}.
For the finite-candidate result, assume uniform approximation errors
$\delta_O,\delta_Q$ for $\widehat{\mathcal O},\widehat{\mathcal Q}$ on
$\mathcal P$, and Lipschitz constants $L_O,L_Q$ for the reference maps
in a fixed parameter norm.
\begin{proposition}[An information limit for forecasting]
\label{S-prop:forecast-diameter}
Suppose $\mathcal C_\varepsilon(y)$ is nonempty and has finite forecast
diameter. For any forecast $a$ based on these observations,
\begin{equation}
 \sup_{p\in\mathcal C_\varepsilon(y)}\|a-\mathcal Q(p)\|
 \ge \tfrac12 D_\varepsilon(y).
 \label{S-eq:forecast-lower-bound}
\end{equation}
Choosing the forecast of any compatible $p_0$ gives worst-case error at most
$D_\varepsilon(y)$. In particular, all compatible explanations have the same
future exactly when $D_\varepsilon(y)=0$, whether or not $p$ is unique.
\end{proposition}
\begin{proof}
For any compatible $p,q$, the triangle inequality gives
$\|\mathcal Q(p)-\mathcal Q(q)\|
\le\|\mathcal Q(p)-a\|+\|a-\mathcal Q(q)\|$.
Taking suprema proves the lower bound. The upper bound follows from the
definition of diameter.
\end{proof}

\begin{proposition}[A finite-candidate forecast bound]
\label{S-prop:candidate-forecast}
Suppose $y=\mathcal O(p_*)+e$ with $\|e\|\le\varepsilon$, and a candidate
$p_b$ lies within distance $h$ of $p_*$. Set
$\tau=\varepsilon+\delta_O+L_Oh$ and
\begin{equation}
 A=\{k:\|\widehat{\mathcal O}(p_k)-y\|\le\tau\}.
\end{equation}
Let $w_k\ge0$ sum to one, with positive total weight on $A$, and let
$\alpha=\sum_{k\notin A}w_k$. Denote by $D_A$ and $D_B$ the diameters of
the approximate future curves in $A$ and in the full candidate pool. Then
the weighted forecast $\bar Q=\sum_k w_k\widehat{\mathcal Q}(p_k)$ satisfies
\begin{equation}
 \|\bar Q-\mathcal Q(p_*)\|
 \le
 \underbrace{(1-\alpha)D_A}_{\text{accepted-curve disagreement}}
 +\underbrace{\alpha D_B}_{\text{weight outside acceptance}}
 +\underbrace{\delta_Q+L_Qh}_{\text{model and candidate error}}.
 \label{S-eq:candidate-forecast-bound}
\end{equation}
\end{proposition}
\begin{proof}
The observation error, approximation error and Lipschitz bound give
$\|\widehat{\mathcal O}(p_b)-y\|\le\varepsilon+\delta_O+L_Oh$, so $b\in A$.
The triangle inequality gives
\[
 \|\bar Q-\widehat{\mathcal Q}(p_b)\|
 \le \sum_{k\in A}w_kD_A+\sum_{k\notin A}w_kD_B
 =(1-\alpha)D_A+\alpha D_B.
\]
Finally,
$\|\widehat{\mathcal Q}(p_b)-\mathcal Q(p_*)\|\le\delta_Q+L_Qh$.
Adding these bounds proves the claim, including $\alpha=0$.
\end{proof}

For weights summing to one and $\bar Q=\sum_k w_kQ_k$, expansion gives
\begin{equation}
 V_Q=\sum_k w_k\|Q_k-\bar Q\|_{\rm RMS}^2
 =\frac{\operatorname{tr}\operatorname{Cov}_w(Q)}{n_F}
 =\frac12\sum_{k,l}w_kw_l\|Q_k-Q_l\|_{\rm RMS}^2.
 \label{S-eq:weighted-disagreement}
\end{equation}
This is an average disagreement, not a compatible-set diameter or error
against truth. If a top-weight set $A$ discards mass
$\delta=\sum_{k\notin A}w_k<1$, the full mean of a bounded scalar $f$ is
$(1-\delta)\mathbb E[f\mid A]+\delta\mathbb E[f\mid A^c]$, giving
$|\mathbb E f-\mathbb E[f\mid A]|\le\delta(\max f-\min f)$.
This controls removal from an existing pool, not alternatives missing from it.
Uniform surrogate errors and candidate coverage are unavailable for the
stored coagulation corpus, so the finite-candidate bound is not a
certificate for its predictions.

\subsection{Linearized ambiguity and prospective measurement}
\label{S-sec:identifiability}
For observation and forecast Jacobians $H_M,G_M$ with respect to missing
inputs, scale each input by its training range and each size by the
observation tolerance. Let $N$ have orthonormal columns spanning
$\ker H_M$.
\begin{proposition}[Linearized forecast ambiguity]
Under the linearized surrogate map, every parameter perturbation invisible to the observations is also invisible to the forecast if and only if
\begin{equation}
 \ker H_M\subseteq\ker G_M.
\end{equation}
Equivalently, let the columns of $N$ form an orthonormal basis of $\ker H_M$, so $N^\top N=I$. Then this condition is $G_MN=0$.
\end{proposition}
\begin{proof}
An observation-invisible perturbation has the form $v=Nc$. Its forecast perturbation is $G_MNc$, which vanishes for every $c$ exactly when $G_MN=0$.
\end{proof}

\begin{proposition}[Forecast value of an affine measurement update]
\label{S-prop:affine-assay}
Let $Q$ be the vector of candidate future sizes and $X=\eta_j$ a candidate factor value, under any working distribution with finite second moments and $\sigma_X^2=\operatorname{Var}(X)>0$. Write $c=\operatorname{Cov}(Q,X)$. Among affine updates of the form $m(X)=\mathbb E Q+b(X-\mathbb E X)$, the minimum expected mean squared forecast error is
\begin{equation}
 \min_b\frac{\mathbb E\|Q-m(X)\|_2^2}{|F|}
 =V_Q-\frac{\|c\|_2^2}{|F|\sigma_X^2},
 \qquad b=\frac{c}{\sigma_X^2}.
 \label{S-eq:affine-assay-risk}
\end{equation}
Thus Eq.~\eqref{S-eq:acquisition} is exactly the reduction in this optimal affine-update risk.
\end{proposition}
\begin{proof}
Expand the numerator as $\operatorname{tr}\operatorname{Cov}(Q)-2b^Tc+\sigma_X^2\|b\|_2^2$ and complete the square in $b$.
\end{proof}

The forecast-measurement score referenced in the affine-update result is
\begin{equation}
 A_j^{\rm forecast}=\frac{\|\operatorname{Cov}(Q,\eta_j)\|_2^2}
 {n_F\operatorname{Var}(\eta_j)}.
 \label{S-eq:acquisition}
\end{equation}
In the affine-update result, $V_Q$ is defined above. Applying the
conditional-mean least-squares property shows that its utility is a lower
bound on the ideal nonlinear conditional-mean risk reduction under the
same distribution; equality holds for jointly Gaussian $(Q,X)$.
A symmetric $X$ and $Q=X^2$ give zero covariance despite perfect information
about $Q$ after observing $X$, so affine utilities need not rank nonlinear
measurement value correctly \cite{chaloner1995design,attia2018goal}.
Replacing $Q$ by the range-scaled parameter vector instead targets
parameter uncertainty, a different objective.

The POD--RBF diagnostic evaluates Jacobians at the weighted input mean
using finite-difference steps of 0.1\% of each training range, clipped at
the bounds. Singular values at or below $10^{-3}$ times the largest are
classified as unresolved. The corresponding right singular vectors form
orthonormal $N$; the forecast gain is $\|G_MN\|_2$, or zero for an empty
unresolved subspace. This thresholded surrogate calculation is distinct
from a global identifiability result.

For a noisy prospective assay $A=X+e$, assume $e$ is independent of
$(Q,X)$, centered, and has variance $r^2$. Then
$\operatorname{Cov}(Q,A)=c$ and $\operatorname{Var}(A)=\sigma_X^2+r^2$.
The same square completion gives the affine forecast-risk reduction
\begin{equation}
 \frac{\|c\|_2^2}{n_F(\sigma_X^2+r^2)},\qquad \sigma_X^2+r^2>0.
 \label{S-eq:noisy-assay-utility}
\end{equation}
If both variances vanish, the covariance and utility are zero. The code
uses a denominator floor $10^{-12}$, which damps the score below that
scale. Independence and finite second moments suffice for this risk
identity; Gaussianity is needed only for the specific likelihood below.
The practical method chooses an assay using the score, then reruns
nonlinear candidate inference. Its realized error is not guaranteed to
equal the affine expected risk.
The exact-assay comparison starts with all five inputs unknown, reveals
one of Fg, IX, VIII or V, and keeps tissue factor unknown. Forecast-directed,
parameter-directed, random and fixed-IX rules each use two pools of 2,048
candidates; the no-assay reference uses one pool. The chosen factor is
revealed only after selection, and each rule has separate interval scaling.

\subsection{Conditioning on uncertain biochemical measurements}
\label{S-sec:noisy-assay-theory}
For range-scaled factors $\eta_j\in[0,1]$, a measured subset $J$ has
values $a_j=\eta_j+e_j$ with independent Gaussian errors of known
standard deviation $r_j>0$. Errors are independent of the inputs and
early-size errors. The likelihood is
\begin{equation}
 \mathcal L_a(\eta)\propto
 \exp\!\left[-\frac12\sum_{j\in J}(a_j-\eta_j)^2/r_j^2\right].
 \label{S-eq:noisy-assay-likelihood}
\end{equation}
Under the independent uniform-box prior, conditioning gives independent
normal densities centered at $a_j$, with scale $r_j$, truncated to $[0,1]$;
unmeasured inputs remain uniform. Inverse distribution functions transform
scrambled Sobol coordinates. Only the early-size likelihood is then used
for weighting, since the assay likelihood has entered the proposal already.
SNODE latent vectors are drawn from the conditional prior for each input.
Reported assays outside the bounds are not clipped, although the
parameter prior excludes out-of-box values. Correlated priors or assay
errors require a different factorization. A value $r_j=0.05$ means five
percent of the training range, not a coefficient of variation or a
measured laboratory precision.

\section{Growth-law examples and complete identification proofs}
\label{S-sec:relaxation-theory}
\subsection{Relaxation of growth speed}
The effective law
\begin{equation}
 \dot L=v+z,\quad \dot z=-\lambda z,\quad
 v,z_0,\lambda\ge0,\quad L(t_0)=L_0>0,\quad z(t_0)=z_0
 \label{S-eq:relaxation-system}
\end{equation}
models a sustained speed plus a temporary contribution. Its solution is
\begin{equation}
 L(t)=L_0+v\tau+z_0G_\lambda(\tau),\qquad
 G_\lambda(\tau)=
 \begin{cases}(1-e^{-\lambda\tau})/\lambda,&\lambda>0,\\
 \tau,&\lambda=0,\end{cases}\quad \tau=t-t_0.
 \label{S-eq:relaxation-solution}
\end{equation}
It is nonnegative and nondecreasing, with a nonincreasing speed.
The hypothesis is motivated by distinctions between initiation and
propagation and observations of coagulation waves
\cite{ovanesov2005initiation,dashkevich2012thrombin,kuprash2018sensitivity};
it is not derived from the unavailable simulator or identified with a
measured biochemical species. It excludes activation bursts, changing
flow and dissolution.
\begin{proposition}[Identification of a curved effective growth law]
\label{S-prop:relaxation-identification}
Suppose four exact size observations at arbitrary strictly increasing times
admit a solution of Eq.~\eqref{S-eq:relaxation-system} with $z_0>0$ and
$\lambda>0$. Then its $(v,z_0,\lambda)$ are unique. The first observation fixes
$L_0$. If the observations instead lie on a nondecreasing line with $L_0>0$,
the output growth law is unique but redundant relaxation parameters need not
be identifiable.
\end{proposition}
\begin{proof}
For $\lambda>0$, write the curve as $a+v\tau-Ae^{-\lambda\tau}$, with
$A=z_0/\lambda>0$ and $a=L_0+A$. If two curved fits have distinct rates, their
difference has four zeros. Rolle's theorem applied twice requires at least two
distinct zeros of its second derivative. That derivative is a nontrivial
difference of two exponentials with distinct rates and positive coefficients,
and can have at most one zero, a contradiction. For equal rates, a nonzero
second derivative has constant sign, also impossible; the exponential
coefficients must agree, followed by the linear coefficients. A curved fit
and a linear fit cannot share four points by the same argument. On the linear
branch only the total constant speed is observable: when $\lambda=0$ it is
$v+z_0$, and when $z_0=0$ the value of $\lambda$ does not affect the output.
\end{proof}

For equally spaced visits, put $d_i=L_{i+1}-L_i$, $i=0,1,2$. Then
\begin{equation}
 q=\frac{d_1-d_2}{d_0-d_1},\quad
 \lambda=-\log(q)/h,\quad
 v=\frac{d_1-qd_0}{h(1-q)},\quad
 z_0=\frac{\lambda(d_0-vh)}{1-q}.
 \label{S-eq:relaxation-explicit}
\end{equation}
A curved admissible solution requires $0<q<1$, $v\ge0$, $z_0>0$.
These are classical Prony-type formulas, not a new fitting principle
\cite{potts2010prony}. In expanded coordinates $z_0=\kappa a_0$,
$(\kappa,a_0)\mapsto(c\kappa,a_0/c)$ preserves every size observation
for any $c>0$. Thus four exact admissible curved observations can give
$D_0=0$ without identifying these multiplicative causes; these causes
are not identified with the five biochemical inputs.

For a perturbed law $\dot L=v+z+r_L$, $\dot z=-\lambda z+r_z$ with
the same $\lambda$, variation of constants followed by reversing the
order of integration gives, for $T\ge s$,
\begin{align}
 |e_L(T)|\le{}&|e_L(s)|+|\Delta v|(T-s)
 +|e_z(s)|G_\lambda(T-s)+\int_s^T|r_L(t)|\,dt \notag\\
 &+\int_s^TG_\lambda(T-t)|r_z(t)|\,dt.
 \label{S-eq:relaxation-discrepancy}
\end{align}
Indeed $e_z(t)=e^{-\lambda(t-s)}e_z(s)+
\int_s^te^{-\lambda(t-u)}r_z(u)\,du$; substitution into $\dot e_L$
and the triangle inequality give the displayed bound. Differences in
decay rate can enter $r_z$. For $\lambda>0$, initial drive error contributes
at most $|e_z(s)|/\lambda$, whereas sustained-speed error grows linearly.
Sparse measurements do not bound these unseen residuals.

Fitting fixes $L_0$ to the first observed size. For each rate trial,
nonnegative least squares fits $v,z_0$ to the other three observations.
The search uses 81 logarithmically spaced rates and bounded refinement on
$[10^{-5}/(t_3-t_0),25/\min_i(t_{i+1}-t_i)]$, with the exact linear
limit included separately. The bounds are numerical, not biochemical.
They need not contain every plausible fit. The reduced model families and
excluded parameter range were selected after inspecting the corpus, so
their empirical comparisons are retrospective.

\subsection{Parallel transport and local growth}
\label{S-sec:front-theory}
A maintained reservoir and quasi-steady linear concentration profile
supply flux $D\Delta C/L$. If a parallel process supplies constant flux
$J$ and each swept volume requires material density $\rho$, then
\begin{equation}
 \dot L=c+\beta/L,\qquad c=J/\rho\ge0,\quad
 \beta=D\Delta C/\rho\ge0,\quad L>0.
 \label{S-eq:front-law}
\end{equation}
For a length $L$, $c$ is length/time and $\beta$ is length$^2$/time.
This is a parallel-process closure; processes in series give a different
law. Its concavity $\ddot L=-\beta\dot L/L^2\le0$ excludes sustained
acceleration. More generally consider $\dot x=af(x)+bg(x)$, where
$a,b\ge0$, $f,g$ are known positive continuous functions on the size
interval, and $g/f$ is strictly increasing.
\begin{proposition}[Three-snapshot identification]
\label{S-prop:three-snapshot}
Let $t_0<t_1<t_2$ and $x_0<x_1<x_2$, with the assumptions on $f,g$ holding on the closed interval $[x_0,x_2]$. Define
\[
 I_{uv}(r)=\int_u^v\frac{dx}{f(x)+r g(x)},\qquad
 R(r)=\frac{I_{x_1x_2}(r)}{I_{x_0x_1}(r)}.
\]
Then $R$ is continuous and strictly decreasing on $[0,\infty)$, with
\[
 \begin{aligned}
 R(0)&=\frac{\int_{x_1}^{x_2}\frac{1}{f(x)}\,dx}{\int_{x_0}^{x_1}\frac{1}{f(x)}\,dx},\\
 R_\infty&:=\lim_{r\to\infty}R(r)
 =\frac{\int_{x_1}^{x_2}\frac{1}{g(x)}\,dx}{\int_{x_0}^{x_1}\frac{1}{g(x)}\,dx}.
 \end{aligned}
\]
There is a unique positive pair $(a,b)$ matching all three observations if and
only if $(t_2-t_1)/(t_1-t_0)$ lies strictly between these limits. Equality at a
limit identifies the corresponding pure process. Ratios outside the closed
interval are incompatible with the model. Two increasing observations instead
permit a continuum of mixtures, each with its own uniquely determined scale.
\end{proposition}

\begin{proof}
For $a>0$, set $r=b/a$. Separation of variables gives
$t_1-t_0=I_{x_0x_1}(r)/a$ and $t_2-t_1=I_{x_1x_2}(r)/a$.
On an interval $A$, let $E_{A,r}$ denote expectation under the density
proportional to $[f(x)+rg(x)]^{-1}$. Differentiating the positive integral yields
\[
 \frac{d}{dr}\log I_A(r)
 =-E_{A,r}\!\left[\frac{g(x)}{f(x)+rg(x)}\right].
\]
The integrand equals $h(x)/(1+rh(x))$ for the strictly increasing function
$h=g/f$, so its expectation on $(x_1,x_2)$ strictly exceeds that on $(x_0,x_1)$.
Thus $d\log R/dr<0$. Continuity and the limits follow from integration on compact
intervals with positive $f,g$; multiplying $I_A$ by $r$ gives the limit involving
$1/g(x)$. The intermediate value theorem identifies $r$ uniquely, then
$a=I_{x_0x_1}(r)/(t_1-t_0)$ and $b=ra$. The limiting pure $g$ case has $a=0$.
With only the first interval, every $r\ge0$ is admissible after this scale change.
\end{proof}

\begin{proposition}[Model-conditional forecast envelope]
\label{S-prop:front-envelope}
For two exact observations $0<L_0<L_1$ at $t_0<t_1$, every solution of
Eq.~\eqref{S-eq:front-law} passing through both satisfies, at $T\ge t_1$,
\begin{equation}
 \sqrt{L_1^2+h(L_1^2-L_0^2)}
 \ \le L(T)\le\ L_1+h(L_1-L_0),\qquad
 h=\frac{T-t_1}{t_1-t_0}.
 \label{S-eq:front-envelope}
\end{equation}
Both bounds are attained: the lower by $c=0$, and the upper by $\beta=0$.
\end{proposition}

\begin{proof}
For any proposed future size $X>L_1$, Proposition~\ref{S-prop:three-snapshot}
shows that $I_{L_1X}(r)/I_{L_0L_1}(r)$ strictly decreases with $r=c/\beta$.
At fixed $r$ it strictly increases with $X$. Consequently the future size
corresponding to the fixed time ratio $h$ increases with $r$. The two limits are
the square-root and linear solutions in Eq.~\eqref{S-eq:front-envelope}; every
intermediate size is attained by continuity. The case $T=t_1$ is immediate.
\end{proof}

\begin{proposition}[Nonexpansive propagation of model discrepancy]
\label{S-prop:front-stability}
Let $L,\widehat L>0$ solve
$\dot L=c+\beta/L+r(t)$ and
$\dot{\widehat L}=c+\beta/\widehat L$ on $[s,T]$, with the same nonnegative
coefficients and an integrable residual. Then
\begin{equation}
 |L(T)-\widehat L(T)|
 \le |L(s)-\widehat L(s)|+\int_s^T|r(t)|\,dt.
 \label{S-eq:front-stability}
\end{equation}
\end{proposition}
\begin{proof}
With $e=L-\widehat L$, the difference equation is
$\dot e=-\beta e/(L\widehat L)+r$. Multiplication by $\operatorname{sign}(e)$
gives $d|e|/dt\le|r|$ almost everywhere; integration proves the bound.
\end{proof}

For the transport law, take $f(x)=1/x$, $g(x)=1$, $a=\beta$, $b=c$.
The three-snapshot compatibility condition becomes
\begin{equation}
 \frac{L_2-L_1}{L_1-L_0}\le
 \frac{t_2-t_1}{t_1-t_0}\le
 \frac{L_2^2-L_1^2}{L_1^2-L_0^2}.
 \label{S-eq:front-compatibility}
\end{equation}
For $c,\beta>0$, travel time is
\begin{equation}
 \mathcal T(L_0,L)=\frac{L-L_0}{c}
 -\frac{\beta}{c^2}\log\frac{cL+\beta}{cL_0+\beta},
 \label{S-eq:front-travel}
\end{equation}
with limits $(L^2-L_0^2)/(2\beta)$ for $c=0$ and $(L-L_0)/c$ for
$\beta=0$. The implementation evaluates these limits stably.

The bounds are sharp model-conditional ranges, not probability intervals.
For observation boxes $L_i\in[L_i^-,L_i^+]$ with
$0<L_0^-$ and $L_0^+<L_1^-$, monotonicity in the two observed sizes gives
the sharp outer bounds
\begin{equation}
 \sqrt{(1+h)(L_1^-)^2-h(L_0^+)^2}
 \le L(T)\le (1+h)L_1^+-hL_0^-.
 \label{S-eq:front-box-envelope}
\end{equation}
The indicated corners and pure processes attain both extrema.
Overlapping or nonpositive boxes do not satisfy the assumptions.

\subsection{Attenuation, uniqueness and the role of assumptions}
\label{S-sec:attenuation-theory}
The attenuation law used in the scalar-flow and final-envelope propositions is
\begin{equation}
 \dot L=c+\beta/L-\gamma L,\qquad
 c,\beta,\gamma\ge0,\quad c+\beta>0.
 \label{S-eq:attenuation-law}
\end{equation}
For $\gamma>0$, its equilibrium
$K=(c+\sqrt{c^2+4\gamma\beta})/(2\gamma)$ is stable. Starting from
$0<L_0<K$, the velocity is positive and strictly decreasing in size, so
the solution grows concavely to $K$ and cannot cross it by uniqueness.
The loss term attenuates propagation, not dissolution of formed clot.
For the basis $(1/L,1,-L)$, multiplication by positive $L$ gives a
polynomial of degree at most two, which verifies the zero-counting
hypothesis of the scalar-flow proposition. Thus four increasing snapshots
admit at most one positive-flow coefficient triple, conditional on existence.

\begin{proposition}[Finite-snapshot uniqueness for a scalar flow]
\label{S-prop:chebyshev-flow}
Let $V_\alpha(x)=\sum_{j=1}^k\alpha_j\phi_j(x)$. Suppose that the basis
functions are continuous on $[x_0,x_k]$ and have the
zero-counting property that every coefficient vector with at least one nonzero
entry defines a linear combination with at most $k-1$ distinct zeros there.
Among coefficient vectors satisfying $V_\alpha(x)>0$ for every
$x\in[x_0,x_k]$, the $k+1$ increasing exact snapshots
$(t_i,x_i)_{i=0}^k$ admit at most one coefficient vector.
\end{proposition}
\begin{proof}
Suppose two such admissible vectors $\alpha,\widetilde\alpha$ fit the
observations. Both velocities are continuous and positive on $[x_0,x_k]$,
so their reciprocals are well defined. Their travel times agree on each
of the $k$ disjoint size intervals:
\[
 0=\int_{x_{i-1}}^{x_i}
 \left(\frac{1}{V_\alpha(x)}-\frac{1}{V_{\widetilde\alpha}(x)}\right)dx
 =\int_{x_{i-1}}^{x_i}
 \frac{V_{\widetilde\alpha}(x)-V_\alpha(x)}
      {V_\alpha(x)V_{\widetilde\alpha}(x)}\,dx.
\]
The denominator is positive, so the continuous numerator has a zero in each
open interval, unless it vanishes identically on one interval. The latter
case already gives infinitely many zeros. A nonzero linear combination would
therefore have at least $k$ distinct zeros, contrary to the hypothesis. Linear
independence follows from the same zero-counting property, so the vectors agree.
\end{proof}

With fixed coefficients and additive residual $r$, the difference equation
gives $d|e|/dt\le-\gamma|e|+|r|$, hence
\begin{equation}
 |e(T)|\le e^{-\gamma(T-s)}|e(s)|
 +\int_s^Te^{-\gamma(T-t)}|r(t)|\,dt.
 \label{S-eq:attenuation-stability}
\end{equation}
The unknown residual is not bounded by this inequality alone.

\begin{proposition}[Unknown attenuation changes forecast ambiguity]
\label{S-prop:attenuation-envelope}
Given two positive increasing snapshots, the closure of the possible future
sizes for Eq.~\eqref{S-eq:attenuation-law} is
\begin{equation}
 L_1\le L(T)\le L_1+h(L_1-L_0),\qquad T>t_1.
 \label{S-eq:attenuation-envelope}
\end{equation}
The upper endpoint is attained by constant-speed growth. The lower endpoint
is an infimum, approached by increasingly rapid saturation but not attained
on a strictly positive-flow branch at finite time.
\end{proposition}
\begin{proof}
Growth gives $L(T)>L_1$ and concavity bounds the future slope by the preceding
secant slope, yielding the upper bound. In the subfamily $\beta=0$, any
$K>L_1$ fits the two observations with
\[
 \gamma(K)=\frac{1}{t_1-t_0}\log\frac{K-L_0}{K-L_1},\quad c=\gamma(K)K,
 \qquad L(T)=K-(K-L_1)e^{-\gamma(K)(T-t_1)}.
\]
As $K\downarrow L_1$ the future approaches $L_1$; as $K\to\infty$ it
approaches the constant-speed endpoint. Continuity supplies every intermediate
future size, with the constant-speed endpoint itself supplied by $\gamma=0$.
\end{proof}

Microscopic parameters remain ambiguous: joint rescaling of $J,\rho,D$,
or inverse rescaling of $D,\Delta C$, preserves $c,\beta$.
Likewise writing $\gamma=k_I\chi/\rho$ introduces a product that size
observations cannot separate. These are symmetries of the reduced laws,
not established symmetries of the coagulation simulator.
Before initiation, $L(t)=v(t-\tau)_+$ can match the same zero observations
for all onsets $\tau$ after the final visit, yet have unbounded future
size as $v$ grows when $\tau<T$. This onset-augmented example lies
outside the positive-front assumptions.

\section{Data partitions, models and numerical protocols}
\label{S-sec:data}
\subsection{Corpus and evaluation roles}
The corpus contains 1,000 unique input vectors and unique curves on 121
common times from 0.0008 to 59.9992 minutes. All stored values are finite;
curves are nonnegative and nondecreasing. The corresponding author confirms
that the trajectories were generated using the original code for the spatial
coagulation model of Panteleev et al.\ \cite{panteleev2006spatial}.
In that model, $\sigmatf$ is tissue-factor surface density on the activating
surface. Clot size is a front distance: the experimental definition uses
the half-maximal light-scattering position, and simulated fibrin profiles
provide its model counterpart. Published growth curves use millimetres
and minutes. The source implementation used Watcom C/C++ 10.0 and an
embedded Runge--Kutta--Fehlberg 2(3) method adapted to one-dimensional
reaction--diffusion equations on 0--3\,mm, with boundary fluxes coupling
surface reactions to plasma transport.
The exact code snapshot, mesh, run-specific boundary settings and solver
tolerances for this archive are unavailable in the supplied materials.
Stored sizes span 0--2.98, consistent with the published spatial domain,
but the MAT file does not document the front-extraction or unit-conversion
convention. The numerical unit and scaling of the stored tissue-factor
surface density remain unverified. The reported factor concentration unit
is nM, without unit metadata in the MAT file.

The seven-model tissue-factor protocol uses ten splits, seeds 42--51,
with 800 training, 100 validation and 100 test curves. Checkpoint selection
and interval scaling both use validation curves. Four distinct visits are
sampled uniformly from times before 30 minutes with size above $10^{-6}$.
If fewer visits qualify, all are used; two test appearances with none
are excluded. In the nominal 16-visit sensitivity test, 29 appearances
have fewer than 16 qualifying visits, including those two exclusions.
For each requested count $m$, 200 schedules use seed
$12345+(s-42)+100000m+r$, $r=0,\ldots,199$. NDE and retrieval pools
are reused across schedules; context-dependent pools are regenerated.

The all-mask protocol uses seeds 42--44 and their 800/100/100 roles.
MLP and POD--RBF selection occurs on an 80/20 split inside the 800
training curves, followed by refitting on all 800. SNODE checkpoints
are reused. Four fixed visits are nearest 5, 12, 20 and 27 minutes,
independent of size; all 100 test cases are retained.
The biochemical-noise, repeated-candidate and integration diagnostics
instead leave those 100 checkpoint-validation curves unused.
A permutation with seed $s+91000$ divides each original test set into
50 calibration and 50 evaluation cases. The first 20 evaluation IDs
define the repeated-candidate and integration subset. Roles are disjoint
within each split; recurring curves do not create independent cohorts.
Scaling always uses training data. Training curves span the full
0--60 minute interval, so later-size prediction is not extrapolation
beyond the training time domain.

For the excluded-range test, all 141 curves with $\sigmatf<1000$ are
withheld from fitting and calibration. Each seed splits the other 859
into 659 training, 100 calibration and 100 in-range test cases.
MLP and POD--RBF are refitted; SNODE is excluded because its saved
training sets contain low-TF curves. Candidate search still covers
the excluded input range.

\subsection{Joint candidate inference and interval evaluation}
\label{S-sec:joint}
Unknown coordinates are represented by missing-value markers and varied
jointly in a scrambled Sobol design on the training-range box. Known
factors stay fixed. All-mask inference uses 2,048 candidates and
weights proportional to
$\exp[-\sum_j(u_k(t_j)-y_j)^2/(2s^2)]$ with temperature one,
$s^2=0.02^2+s_{\rm noise}^2$ in stored units. SNODE has one
conditional-prior latent draw per input and no second latent penalty.
If no visits remain, all candidates have equal weights.
The historical tissue-factor calculation instead uses 128 grid values,
32 prior draws per value, latent temperature 1.35, a quadratic latent
penalty 0.05, top-30 retention and weight temperature two.
Context and PCA candidates already depend on observations before
scoring, so these methods do not share one observation-independent prior.

Gaussian size noise has standard deviation 1\% or 5\% of the training
maximum size. Missing-visit tests delete each visit independently with
probability 0.5. Values are not clipped and the target stays uncorrupted.
Each model/mask/setting has validation-scaled pointwise 90\% bands;
a second check reuses the clean-data multiplier under corruption.
Scaling uses the empirical 90th percentile of absolute error divided
by the nominal Gaussian half-width, with spread floors $10^{-6}$ times
the corresponding training size or input range. Whole-curve coverage
requires all sampled future values to be covered. Interval width and
$\mathrm{IS}_{.1}(L,U;y)=U-L+20(L-y)_++20(y-U)_+$ assess sharpness and
misses as well as coverage. These are predictive bands for uncorrupted
simulator outputs, not new noisy measurements.

Pooled relative $L^2$ combines squared errors across cases before
normalization; mean case relative $L^2$ normalizes each case separately.
Parameter MAE is divided by the appropriate training range and averaged
only over missing coordinates. Mask-group summaries give combinations
equal weight after averaging three runs. Overlapping masks, settings and
splits are not independent datasets.

\subsection{Latent neural ordinary differential equations}
\label{S-app:surrogate-details}

For each complete input vector $\theta$, SNODE draws an eight-dimensional
random vector $\xi\sim p_\phi(\xi\mid\theta)$ from a learned Gaussian
distribution. This vector is held fixed while generating one trajectory.
The initial size is positive, and the learned growth rate is
\begin{equation}
 F_\phi=\min\!\left\{50,
 \softplus(r_\phi)\,\frac{\log(1+\exp(5(K_\phi-u)))}{5}\right\},
 \qquad K_\phi=\softplus(k_\phi)+10^{-4}.
\end{equation}
Here $r_\phi$ and $k_\phi$ are network outputs and
$\softplus(x)=\log(1+e^x)$. Positive growth increments keep the computed
trajectory nonnegative and nondecreasing under the implemented two-stage
Heun method. Although $K_\phi$ acts like a capacity parameter, the softplus
factor does not enforce a hard bound $u\le K_\phi$. The model does not enforce
the coagulation reaction equations, conservation laws or simulator residuals.
Its randomness comes from $\xi$, not a Brownian driving process.

Variational training uses a full-trajectory encoder
$q_\phi(\xi\mid\theta,u_{0:T})$ and a conditional prior, with reconstruction
and KL terms \cite{kingma2022autoencodingvariationalbayes}.
Test prediction uses the prior, not this encoder. The precise losses,
checkpoint criteria and sampling temperatures are recorded below.

One curve per complete input vector supplies no repeated measurement of
biological randomness; latent spread is evaluated as prediction uncertainty.

The SNFDE model also uses past generated sizes, at delays of
1, 2, 4, 8, 16, 32 and 60 time-grid steps. It retains the eight-dimensional
latent vector and nondecreasing-growth restriction. Neural delay equations
\cite{zhu2021neural} and functional differential equations
\cite{hale1993fde} provide the modelling framework. The Mori--Zwanzig
representation explains why removing unobserved states can introduce dependence
on the past \cite{chorin2000optimal,gouasmi2017memory}. Here the selected delay
taps are a learned approximation; no memory kernel is derived from the
coagulation equations, and the comparison does not identify biochemical memory.

\subsection{Direct neural and interpolation controls}

The monotone MLP has four hidden layers of width 192 with SiLU activations.
It outputs 121 positive increments, which are accumulated to form the predicted
curve. Training uses AdamW, learning rate $10^{-3}$, weight decay $10^{-5}$ and
batch size 128. The loss is normalized trajectory mean squared error plus
one-half the terminal mean squared error. An inner validation split selects
the epoch count, up to 1,200 epochs, before refitting on all training cases.

POD--RBF combines proper orthogonal decomposition with radial-basis-function
interpolation. It first represents each training curve by 12 dominant curve
patterns, or modes. It then interpolates their
coefficients as functions of the biochemical inputs. Inner training validation
compares thin-plate-spline and cubic radial-basis kernels with smoothing values
$0,10^{-4},10^{-2},10^{-1}$. Reconstructed curves are adjusted to be
nonnegative and nondecreasing. Their candidate spread comes from uncertain inputs; validation scaling also reflects prediction error.

\subsection{Other models and weights in the seven-model experiment}

The seven-method comparison includes SNODE, SNFDE, a latent context
process, a five-member probabilistic monotone ensemble, stored-trajectory
retrieval, PCA--ridge regression and Gompertz-curve retrieval. The context and
ensemble methods build on existing probabilistic approaches
\cite{garnelo2018cnp,lakshminarayanan2017deepensembles}. The simpler controls
use nearest-neighbour regression \cite{altman1992kernelnearest},
principal-component analysis \cite{jolliffe2002pca}, ridge regularization
\cite{hoerl1970ridge} and Gompertz growth curves \cite{gompertz1825mortality}.
PCA--ridge is not Gaussian-process regression. Its Gaussian component describes
errors in the curve coefficients. Its regression takes the complete candidate
input vector together with padded observation times, sizes and masks.

\subsection{Capacity, optimization and reproducibility}
\label{S-app:reproducibility}

Table~\ref{S-tab:repro-capacity} records capacities checked against all
80 saved neural checkpoints. Counts include training-only encoders but
exclude fixed time frequencies. Capacities are not matched.

\begin{table}[!htbp]
\centering
\caption{Model capacity and candidate count per case. Hidden layers use SiLU.
The notation $192\times d$ means $d$ hidden layers of width 192. Counts in
the second column are gradient-trained weights and biases; a dash denotes
a non-neural fitting method, not an unfitted model. Candidate counts are
before aggregation.}
\label{S-tab:repro-capacity}
\begin{tabular}{@{}>{\raggedright\arraybackslash}p{.15\linewidth}r>{\raggedright\arraybackslash}p{.44\linewidth}r@{}}
\toprule
Model & Weights & Architecture or fitted representation & Candidates\\
\midrule
SNODE & 602,723 & Five $192\times4$ networks; latent dimension 8 & 4,096\\
SNFDE & 396,898 & Three $192\times4$ networks and one $192\times2$ network; latent dimension 8 & 4,096\\
Latent context & 299,481 & Three $192\times3$ networks; latent dimension 8 & 512\\
Ensemble & 655,930 & Five $192\times3$ networks; 131,186 weights per member & 640\\
Library & --- & 800 stored training curves & 800\\
PCA--ridge & --- & 12 modes, $54\times12$ ridge coefficients and $12\times12$ residual covariance & 512\\
Gompertz & --- & Three coefficients fitted to each of 800 training curves & 800\\
\midrule
Joint SNODE & 602,723 & Archived SNODE, without retraining & 2,048\\
Monotone MLP & 135,673 & $5\to192\times4\to121$ positive increments & 2,048\\
POD--RBF & --- & 12 modes and 800 interpolation centres & 2,048\\
\bottomrule
\end{tabular}
\end{table}

\subsubsection{Neural optimization and initialization}

The archived neural models use AdamW, learning rate $8\times10^{-4}$,
weight decay $10^{-5}$, batch size 128 and gradient-norm clipping at five.
SNODE and SNFDE allow 1,200 epochs; the context process and each ensemble
member allow 600. Validation is evaluated at epoch one and every 25 epochs.
An improvement must exceed $10^{-8}$; otherwise the stopping counter increases
by 25. Patience is 250 epochs for the differential-equation models and 150
for the other neural models. The selected weights are restored after stopping.
SNFDE alone uses a cosine learning-rate schedule, ending at
$0.08\times8\times10^{-4}$ at epoch 1,200; the other archived learning rates
remain constant. No equal-budget architecture search was recorded.

SNODE minimizes full-curve MSE plus 0.10 times terminal MSE, 0.05 times
increment MSE and 0.02 times full-curve area MSE, all in normalized coordinates.
Its KL coefficient is $0.002\min(1,e/400)$ at epoch $e$. SNFDE adds the same
reconstruction objective evaluated at the conditional-prior mean with weight
0.25. The context model uses the same reconstruction weights and KL maximum,
but a 300-epoch KL warm-up. Ensemble members minimize a Gaussian
negative log-likelihood for normalized size increments. No trajectory-level
latent variable is fitted for the ensemble.

SNODE selects its checkpoint by validation reconstruction using the
full-curve encoder mean. SNFDE selects by validation prediction using the
conditional-prior mean. The context process selects by prediction from its
context-conditioned prior mean, and the ensemble by increment likelihood.
Recorded selected epochs range from 525--1,200 for SNODE, 850--1,200 for
SNFDE, 500--600 for the context model, and 50--175 per ensemble member.
Last training epochs range from 775--1,200 for SNODE, 1,100--1,200 for
SNFDE and 200--325 for ensemble members; every context fit runs 600 epochs.
Selected and final epochs describe different quantities.

Linear layers use the installed PyTorch default initialization except for
the final biases: SNODE initial-size and capacity biases are $-5$ and $0.5$;
SNFDE dynamics and capacity biases are $-2$ and zero; context and ensemble
decoder biases are $-4$. SNODE and the context/ensemble baselines sample
training indices with replacement within each epoch. SNFDE and the monotone
MLP use shuffled passes without replacement. MLP inner selection checks MSE every 20 epochs with patience 160, then
refits for 940, 280 and 720 epochs for seeds 42, 43 and 44. It uses neither a learning-rate scheduler
nor gradient clipping.

\subsubsection{Numerical integration and conditioning}

SNODE uses fixed-step Heun integration and SNFDE uses explicit Euler,
each with one step per consecutive pair of the 121 stored times. Times
are first mapped to $[0,1]$ using training extrema. Thus there are no adaptive
solver tolerances; solver-step convergence has not been established by the
archived comparison. A 32-component fixed sinusoidal embedding is used by
both models. SNODE's encoder takes all 121 training sizes. SNFDE's encoder
takes 16 uniformly indexed sizes and six curve summaries. Neither encoder
is used to infer a held-out curve from sparse visits.

At step $k$, SNFDE delay $d$ reads generated size $\max(0,k-d)$,
repeating the initial value before the delay is available. It cannot read
future observations. SNFDE starts at the stored first size, exactly zero
for all 1,000 curves; SNODE predicts a positive initial size. Capacity,
integration, loss and checkpoint differences prevent isolating delay benefit.

The seven-model context feature vector has dimension 53: five standardized
candidate inputs and 16 padded values each for observation times, sizes and
masks. The context decoder takes this vector, the five candidate inputs and
eight latent coordinates. The ensemble predicts means and scales for all
121 increments. PCA--ridge uses the same 53 context features plus an
intercept, ridge coefficient $10^{-3}$ and eight random contexts per
training curve. Its residual covariance is regularized by $10^{-6}I$ and
eigenvalues are floored at $10^{-8}$. Gompertz fitting uses bounded nonlinear
least squares with at most 8,000 function evaluations per training curve;
the fitted library, including fit-success flags, is archived. These methods
do not have neural latent dimensions or optimizer epochs.

\subsubsection{Randomness, preprocessing and reproducibility limits}

Input means and standard deviations, time extrema and the maximum size used
for normalization are obtained from training data. Archived split and global
initialization seeds are 42--51; observation seeds are 12345--12354. The
baseline module's separate context-sampling generators retain their fixed
seed-42 offsets across runs.

Experiments were executed with Python 3.14.4 and PyTorch 2.11.0+cu130 on an
NVIDIA RTX PRO 6000 Blackwell GPU. The authors' computational archive retains checkpoints, split
manifests, training histories and command logs, but not a complete historical
software lockfile. The accompanying code-only package excludes these
checkpoints, split files and saved numerical outputs. Source, checkpoint and stopping-epoch records accompany the computations. Seed control does
not guarantee bitwise GPU reproducibility. The seven-model workflow took 8.82 elapsed hours, not measured GPU utilization or per-case latency. Missing original simulator code still prevents an audited
simulator speed-up or regeneration claim.

\Needspace{9\baselineskip}
\subsection{Paired comparison and parameter-space audit}
\label{S-app:statistical-comparison}

\subsubsection{What the ten-run comparison supports}

We re-evaluate the original SNODE--SNFDE comparison at the run level, pairing
the models on the same split and observation schedule. This is a targeted,
retrospective check of the seven-model ranking, not a search over
all model pairs. For run $s$, let $d_s$ be the pooled future relative
$L_2$ error of SNFDE minus that of SNODE. Positive differences favour SNODE.
The ten differences give
\begin{equation}
 \bar d=0.000624,\qquad s_d=0.001668,\qquad
 \bar d\pm t_{0.975,9}\frac{s_d}{\sqrt{10}}
 =[-0.000570,\,0.001817].
\end{equation}
Here $s_d$ is the sample standard deviation. Using the $t$ critical value
2.262, rather than the normal value 1.96, accounts for estimating the variance
from ten runs under the working normal-difference model. It does not establish
normality or remove the limitations of repeated evaluation on one corpus.

The paired statistic is $t=1.182$ with nine degrees of freedom and a
two-sided reference $p=0.267$. The standardized paired difference is
$d_z=\bar d/s_d=0.374$; SNODE has lower error in seven of ten runs. As a
distributional sensitivity check, we enumerate all $2^{10}=1,024$ sign
patterns and compare the absolute signed mean with $|\bar d|$. There are
260 equally or more extreme patterns, giving $p=0.254$. Enumeration is
exact for the stated sign-symmetry reference; it is not a randomized
treatment experiment. Validity of this reference requires joint invariance
to sign flips under the null.

Both calculations are conditional, descriptive comparisons of the archived
computational procedure. A paired $t$ interpretation assumes independent,
approximately normal run differences conditional on the fixed corpus.
Random partitions can assess variability of that procedure, but their reuse
of trajectories does not create independent biological cohorts. The ten
splits have 1,000 test appearances from 665 distinct curves; after eligibility
filtering, the predictions cover 998 appearances from 663 distinct curves.
Appearances are not treated as independent subjects. Neither reference test supports
the claim that SNODE is superior to SNFDE. Conversely, lack of rejection
does not prove equivalence: no equivalence margin was specified. The
comparison also retains the unequal model and inference settings detailed
in Section~\ref{S-app:reproducibility}.

\subsubsection{Observed sampling structure and train--test proximity}

The stored inputs contain identifiable design structure even though the
generation script is missing. Rows 1--243 contain exactly the complete
$3^5$ factorial combination of three levels per biochemical input. In the
order $(\sigmatf,\mathrm{Fg},\mathrm{IX},\mathrm{VIII},\mathrm{V})$, those
levels are the lower bounds $(100,4860,49,0.36,11.43)$, the interior values
$(1000,7600,90,0.70,20)$, and the upper bounds $(10000,10310,130.91,1.04,28.57)$. The other 757 rows are
distinct off-grid input vectors; none of their coordinates equals a
factorial level. This is a check of the stored matrix, not a reconstruction
of its sampling algorithm. The entire corpus should therefore not be
described as 1,000 independent uniform draws. Across all rows, the largest
absolute pairwise Pearson and Spearman correlations are 0.0305 and 0.0372,
respectively. Small pairwise correlations do not establish independence
or a uniform sampling distribution.

For each of the 1,000 test appearances, we compute the Euclidean distance
to the nearest training input after dividing each coordinate difference
by its full stored range. The median distance is 0.195, with fifth and
95th percentiles 0.091 and 0.323; the minimum is 0.059. All test inputs lie
within their split's training coordinate-wise ranges, and no complete
input vector is shared between training and test within a split. These
checks characterize a same-corpus parameter-space comparison. Coordinate-wise
range membership does not establish convex-hull membership, and proximity
alone is not data leakage. Nearest-neighbour IDs and distances are retained for reproduction.

\section{Complete comparison and sensitivity results}
\subsection{Seven-model tissue-factor inference}
\begin{table}[htbp]
\centering
\small
\setlength{\tabcolsep}{4pt}
\caption{Seven-model one-unknown comparison, checked from saved predictions and averaged over ten runs. Both coverage columns use calibrated 90\% bands. Width and interval score are in stored size units; lower score is better. Candidate counts differ by method.}
\label{S-tab:archive}
\begin{tabular}{lrrrrr}
\toprule
 & & \multicolumn{2}{c}{Coverage} & & \\
\cmidrule(lr){3-4}
Method & Pooled $L^2$ & Pointwise & Whole curve & Width & Int. score \\
\midrule
SNODE & 0.01397 & 0.907 & 0.471 & 0.063 & 0.084 \\
SNFDE & 0.01459 & 0.900 & 0.436 & 0.072 & 0.098 \\
Latent context & 0.01667 & 0.905 & 0.702 & 0.099 & 0.136 \\
Deep ensemble & 0.02402 & 0.914 & 0.824 & 0.123 & 0.197 \\
Library & 0.04797 & 0.900 & 0.831 & 0.167 & 0.266 \\
PCA--ridge & 0.06413 & 0.913 & 0.860 & 0.322 & 0.427 \\
Gompertz & 0.15650 & 0.913 & 0.766 & 0.939 & 1.278 \\
\bottomrule
\end{tabular}
\end{table}
The paired comparison above does not establish SNODE superiority to SNFDE.
Their similar errors coexist with differing capacity, integration,
optimization, initialization and candidate budgets.

\subsection{Growth-law adequacy and training-free controls}
Each growth law passes 100 constructed-case checks against an independent
adaptive ODE solver, with largest forward differences below $10^{-8}$.
For three primary visits, the transport law matches 304 sequences;
604 slow down too strongly and 92 contain a nonpositive observation.
Among 918 cases eligible for its two-visit envelope, pointwise and
whole-curve inclusion are 34.3\% and 11.1\%. These are model-conditional
bounds, not nominal probability intervals.

Four-visit comparison requires positive, strictly increasing early sizes:
894 cases qualify, 92 have nonpositive values and 14 have repeats.
Attenuation gives 561 numerical exact fits; relaxation gives 798 exact
and 96 approximate fits, with three rate-boundary solutions.
A numerical exact fit uses maximum relative interval-time error
$10^{-7}$ for attenuation, or maximum size error
$10^{-7}\max(1,\max_i y_i)$ for relaxation. Approximate fits are retained.
\begin{table}[htbp]
\centering
\caption{Simple-model forecast errors with clean observations. Each column uses the same 894 eligible cases for all methods, but case membership may differ between columns. Fits use early sizes only, with no biochemical inputs or training on other curves. Primary visits are near 5, 12, 20, 27 minutes; later visits are near 12, 17, 22, 27. Both predict 30--60 minutes.}
\label{S-tab:front-models}
\begin{tabular}{lrr}
\toprule
Model & Primary pooled $L^2$ & Later pooled $L^2$ \\
\midrule
Linear extrapolation & 0.07068 & 0.06403 \\
Pure-diffusion extrapolation & 0.08648 & 0.08417 \\
First-order saturation & 0.13939 & 0.10425 \\
Transport with attenuation & 0.08368 & 0.08204 \\
Growth-speed relaxation & 0.05034 & 0.06815 \\
\bottomrule
\end{tabular}
\end{table}

\begin{table}[htbp]
\centering
\caption{Reference comparison with all five biochemical inputs unknown; means over three test splits, retaining all 100 cases per split. Surrogates learn from other complete curves and use 2,048 candidates. Relaxation fits four early sizes without cross-case training. The comparison is not matched in training information or computing cost.}
\label{S-tab:front-surrogate}
\begin{tabular}{lrr}
\toprule
Method & Pooled relative $L^2$ & Mean case relative $L^2$ \\
\midrule
SNODE & 0.03698 & 0.04781 \\
Monotone MLP & 0.03713 & 0.04638 \\
POD--RBF & 0.04640 & 0.05043 \\
Relaxation with linear fallback & 0.05662 & 0.06977 \\
Linear extrapolation & 0.07646 & 0.10158 \\
\bottomrule
\end{tabular}
\end{table}
With uniform errors in $\pm0.01$ stored units, 912 cases qualify and
pooled errors are 0.07321 for relaxation and 0.07386 for a line.
At $\pm0.05$, 898 qualify and errors are 0.12869 and 0.11208.
These absolute uniform errors differ from the Gaussian surrogate tests.
The all-case rule uses relaxation for eligible observations and linear
extrapolation otherwise; its choice depends only on early observations.
The four-visit theorem does not apply to fallback cases.

\subsection{All 31 missing-input combinations}
We evaluated all 31 missing-input combinations with three models, three data splits and 2,048 candidates per case. Each split has 100 test and 100 validation cases. The groups with one through five missing inputs contain 5, 10, 10, 5 and 1 combinations, respectively. Table~\ref{S-tab:joint} averages run-level pooled errors, giving each combination equal weight within its group.

Across all combinations, the forward monotone MLP has mean pooled error 0.02785, compared with 0.03130 for SNODE and 0.04843 for POD--RBF. The MLP is more accurate than SNODE in 67 of the 93 paired run--mask comparisons. These reused-data comparisons are not 93 independent experiments and do not establish a general benefit from differential-equation integration.

\begin{table}[htbp]
\centering
\small
\setlength{\tabcolsep}{4pt}
\caption{Joint estimation with 2,048 candidates per case. Values average three runs and all combinations with the stated number of unknowns. Parameter error is range-normalized MAE; coverage is pointwise after validation scaling. ESS measures effective candidate count.}
\label{S-tab:joint}
\begin{tabular}{rlrrrr}
\toprule
Missing & Method & Pooled $L^2$ & Param. error & Coverage & ESS \\
\midrule
1 & SNODE & 0.0218 & 0.062 & 0.905 & 463.6 \\
1 & Monotone MLP & 0.0190 & 0.052 & 0.908 & 487.9 \\
1 & POD--RBF & 0.0649 & 0.138 & 0.889 & 402.3 \\
\addlinespace[4pt]
2 & SNODE & 0.0295 & 0.137 & 0.903 & 244.7 \\
2 & Monotone MLP & 0.0254 & 0.134 & 0.912 & 249.3 \\
2 & POD--RBF & 0.0500 & 0.164 & 0.894 & 200.3 \\
\addlinespace[4pt]
3 & SNODE & 0.0341 & 0.168 & 0.919 & 160.1 \\
3 & Monotone MLP & 0.0304 & 0.167 & 0.912 & 161.1 \\
3 & POD--RBF & 0.0416 & 0.179 & 0.914 & 133.8 \\
\addlinespace[4pt]
4 & SNODE & 0.0377 & 0.186 & 0.923 & 109.0 \\
4 & Monotone MLP & 0.0345 & 0.187 & 0.920 & 108.3 \\
4 & POD--RBF & 0.0429 & 0.192 & 0.921 & 95.3 \\
\addlinespace[4pt]
5 & SNODE & 0.0370 & 0.198 & 0.914 & 73.4 \\
5 & Monotone MLP & 0.0371 & 0.200 & 0.925 & 72.7 \\
5 & POD--RBF & 0.0464 & 0.203 & 0.917 & 69.1 \\
\bottomrule
\end{tabular}
\end{table}
Which input is missing also matters. With one unknown, SNODE's range-normalized parameter error is largest for fibrinogen (0.126) and smallest for $\sigmatf$ (0.023). Averages by missing count hide this difference. The ordering depends on the stored ranges and visit times.

The POD--RBF sensitivity check finds unresolved input directions in 15.3\% of four-unknown cases and all five-unknown cases, using the stated singular-value threshold at the estimated input mean. For five unknowns, the mean forecast-effect measure $\|G_MN\|_2$ is 26.43. Thus input changes that are hard to detect early can still affect this surrogate's future prediction. The check does not establish the same ambiguity in the mechanistic simulator.

\subsection{Size noise, missing visits and weighting}
Noise and missing visits increase error for both neural surrogates (Table~\ref{S-tab:robustness}). The noise percentage is a standard deviation relative to the training-set maximum size, not to each observation. Recalibration for the corrupted setting often brings average pointwise coverage near 90\%, but does not recover clean-data accuracy or ensure whole-curve coverage. Keeping the clean-data multiplier tests what happens without recalibration.

\begin{table}[htbp]
\centering
\small
\setlength{\tabcolsep}{4pt}
\caption{Noise and missing-visit tests, averaged over all masks and three runs. Matched coverage uses calibration for the same observation setting; clean coverage reuses the clean-data multiplier. Both are pointwise, not whole-curve or subgroup guarantees.}
\label{S-tab:robustness}
\begin{tabular}{llrrrr}
\toprule
Setting & Method & Pooled $L^2$ & Cov. matched & Cov. clean & Int. score \\
\midrule
Clean & SNODE & 0.0313 & 0.912 & 0.912 & 0.213 \\
Clean & Monotone MLP & 0.0278 & 0.913 & 0.913 & 0.223 \\
Clean & POD--RBF & 0.0484 & 0.905 & 0.905 & 0.486 \\
\addlinespace[4pt]
1\% noise & SNODE & 0.0546 & 0.902 & 0.873 & 0.318 \\
1\% noise & Monotone MLP & 0.0534 & 0.902 & 0.877 & 0.319 \\
1\% noise & POD--RBF & 0.0682 & 0.902 & 0.917 & 0.491 \\
\addlinespace[4pt]
5\% noise & SNODE & 0.1248 & 0.901 & 0.862 & 0.722 \\
5\% noise & Monotone MLP & 0.1246 & 0.900 & 0.879 & 0.720 \\
5\% noise & POD--RBF & 0.1302 & 0.898 & 0.944 & 0.778 \\
\addlinespace[4pt]
50\% missing & SNODE & 0.0860 & 0.912 & 0.916 & 0.404 \\
50\% missing & Monotone MLP & 0.0846 & 0.907 & 0.914 & 0.419 \\
50\% missing & POD--RBF & 0.1043 & 0.908 & 0.925 & 0.678 \\
\bottomrule
\end{tabular}
\end{table}
We also test keeping only the 30 highest-weight candidates when the unknown inputs are TF, TF with Fg, TF with Fg and IX, or all five. This removes an average of 50.8\% of SNODE's candidate weight. Its mean pooled error changes from 0.02571 to 0.02410. Discarded weight describes the distributional change within the sampled set.

\begin{table}[htbp]
\centering
\small
\setlength{\tabcolsep}{4pt}
\caption{Effect of candidate count on pooled forecast error for seed 42. SNODE's latent draws also change with the budget. This is a sampling-sensitivity check, not proof that the calculation has converged.}
\label{S-tab:budget}
\begin{tabular}{lrrrr}
\toprule
 & \multicolumn{2}{c}{SNODE} & \multicolumn{2}{c}{Monotone MLP} \\
\cmidrule(lr){2-3}\cmidrule(lr){4-5}
Missing inputs & 512 & 2,048 & 512 & 2,048 \\
\midrule
TF & 0.01263 & 0.01301 & 0.01186 & 0.01184 \\
TF,Fg & 0.02097 & 0.02014 & 0.02030 & 0.01631 \\
TF,Fg,IX & 0.04065 & 0.03035 & 0.04119 & 0.03334 \\
All five & 0.05068 & 0.04037 & 0.05285 & 0.03790 \\
\bottomrule
\end{tabular}
\end{table}
The 512-versus-2,048 check changes SNODE latent draws as well as the input pool; the nested comparison below fixes prefix trajectories.

\subsection{Excluding a tissue-factor range and revealing one factor}
Both refitted controls perform much worse in the excluded low-tissue-factor range (Table~\ref{S-tab:holdout}). For tissue factor alone unknown, the MLP's mean pooled error rises from 0.00626 in the in-range control set to 0.41896 in the holdout; pointwise coverage falls to 0.169. Calibration within the training range does not transfer to this range for either control. The groups also differ in curve size and onset time, so this test does not identify a single cause of failure. No uncontaminated NDE comparison is available for this holdout.

\begin{table}[htbp]
\centering
\small
\setlength{\tabcolsep}{4pt}
\caption{Prediction outside the training tissue-factor range, averaged over three refits on 659 cases. Each refit has 100 in-range control and 100 calibration cases. The same 141 low-TF test curves are used each time. Coverage is pointwise; saved NDEs are excluded because their training included low-TF cases.}
\label{S-tab:holdout}
\begin{tabular}{llrrrr}
\toprule
 & & \multicolumn{2}{c}{Pooled error} & \multicolumn{2}{c}{Coverage} \\
\cmidrule(lr){3-4}\cmidrule(lr){5-6}
Method & Missing & In range & Low TF & In range & Low TF \\
\midrule
Monotone MLP & TF & 0.0063 & 0.4190 & 0.862 & 0.169 \\
Monotone MLP & TF,Fg & 0.0091 & 0.3721 & 0.867 & 0.182 \\
Monotone MLP & All five & 0.0262 & 0.4548 & 0.906 & 0.242 \\
\addlinespace[4pt]
POD--RBF & TF & 0.0139 & 0.4952 & 0.871 & 0.127 \\
POD--RBF & TF,Fg & 0.0125 & 0.4409 & 0.869 & 0.156 \\
POD--RBF & All five & 0.0279 & 0.4703 & 0.903 & 0.180 \\
\bottomrule
\end{tabular}
\end{table}
\subsection{Choosing an additional biochemical measurement}

Adaptive rules choose a factor before its value is revealed; tissue factor is never measured. Random and fixed-IX controls use the same assay budget (Table~\ref{S-tab:assay}).

\begin{table}[htbp]
\centering
\small
\setlength{\tabcolsep}{4pt}
\caption{One simulated biochemical measurement after four size observations; means over three runs. All five inputs are initially unknown. Measurement rules have equal assay and candidate budgets; the no-assay reference uses half as many candidates. Coverage is pointwise after rule-specific calibration.}
\label{S-tab:assay}
\begin{tabular}{llrrr}
\toprule
Model & Selection rule & Pooled $L^2$ & Coverage & Int. score \\
\midrule
SNODE & No assay & 0.03698 & 0.914 & 0.349 \\
SNODE & Random factor & 0.03810 & 0.896 & 0.298 \\
SNODE & Fixed IX & 0.03504 & 0.904 & 0.282 \\
SNODE & Input-uncertainty rule & 0.03523 & 0.909 & 0.280 \\
SNODE & Forecast-uncertainty rule & 0.03319 & 0.925 & 0.237 \\
\addlinespace[4pt]
Monotone MLP & No assay & 0.03713 & 0.925 & 0.405 \\
Monotone MLP & Random factor & 0.03773 & 0.909 & 0.323 \\
Monotone MLP & Fixed IX & 0.03608 & 0.909 & 0.318 \\
Monotone MLP & Input-uncertainty rule & 0.03599 & 0.894 & 0.332 \\
Monotone MLP & Forecast-uncertainty rule & 0.03411 & 0.920 & 0.333 \\
\bottomrule
\end{tabular}
\end{table}
For SNODE, the forecast-uncertainty rule gives mean error 0.03319, 12.9\% lower than random factor selection (0.03810); the fixed-IX control gives 0.03504. For Monotone MLP, the forecast-uncertainty rule gives mean error 0.03411, 9.6\% lower than random factor selection (0.03773); the fixed-IX control gives 0.03608. Assays are exact and equal-cost. The MLP forecast rule has a worse interval score than random and fixed-IX controls.

\subsection{Repeated candidate pools and time integration}

\label{S-sec:candidate-stability}
The unchanged MLP uses three scrambled Sobol pools of 32,768 candidates
per case, with prefixes 512, 2,048 and 8,192 sharing their forward values.
The first 20 prespecified evaluation IDs per seed 42--44 give 60 appearances
from 59 curves. All five inputs have independent uniform training-range
priors; four clean sizes use tolerance 0.02. Scrambles are numerical repeats,
not additional cases.

\begin{table}[htbp]
\centering
\small
\setlength{\tabcolsep}{5pt}
\caption{Candidate sensitivity for the MLP only. Pooled error is averaged over the nine seed--scramble groups, each with 20 cases. Forecast change is the mean case relative difference from the corresponding 32,768-point forecast; TF change is divided by the training TF range. Scramble SD is the root mean square of the pointwise sample standard deviations across three scrambles, then averaged over the 60 case appearances. The largest pool is a numerical reference, not a certified converged solution.}
\label{S-tab:candidate-stability}
\begin{tabular}{rrrrr}
\toprule
Candidates & \shortstack{Pooled error\\(\%)} & \shortstack{Forecast change\\(\%)} & \shortstack{TF change\\(\%)} & \shortstack{Scramble SD\\(stored units)} \\
\midrule
512 & 2.91 & 3.49 & 1.46 & 0.01411 \\
2,048 & 2.85 & 2.54 & 0.70 & 0.00724 \\
8,192 & 2.76 & 0.83 & 0.28 & 0.00364 \\
32,768 & 2.74 & -- & -- & 0.00161 \\
\bottomrule
\end{tabular}
\end{table}
Increasing budget reduces average numerical disagreement, not every case's
error. The 2,048-point relative change is 2.54\%; even 32,768-point scrambles
still disagree. Higher effective sample size does not exclude missed
alternatives or model bias.

On the same 20 cases per seed, a separate MLP evaluation at the true biochemical inputs gives mean pooled future error 1.02\%.
This uses information unavailable to inverse inference; its error cannot be subtracted from inverse error as an additive decomposition.

\paragraph{The same check for SNODE.}
SNODE uses the same 60 appearances and biochemical candidate seeds, with
three independently seeded conditional-prior latent pools. Smaller budgets
take joint input--latent prefixes, preserving fitted weights and rollouts.
At 2,048 candidates, mean pooled forecast error is 2.91\%,
and the mean case relative change from its 32,768-candidate reference is 3.57\%.
At 32,768 candidates, mean pooled error is 2.73\%, compared with 2.74\% for the MLP on these cases;
the mean cross-scramble forecast standard deviation remains 0.00237 in stored size units.
This variability combines biochemical candidate sampling and latent sampling;
it does not isolate their separate contributions. The largest pool is a finite numerical reference. MLP uses CPU and SNODE GPU evaluation; equal candidate counts do not match work.

\paragraph{Integration-step sensitivity.} For each of three fixed SNODE checkpoints, we evaluate 20 prespecified held-out inputs with 32 conditional-prior draws each. The biochemical inputs and latent draws are identical across integration rules. Relative to fixed-weight RK4 with four substeps per stored interval, Heun with one, two and four substeps has mean candidate RMS differences of $7.00\times10^{-3}$, $2.01\times10^{-3}$ and $3.73\times10^{-4}$, respectively, after division by the training maximum size; means equally weight the three runs. The largest pointwise difference for one-step Heun is $9.54\times10^{-2}$ stored size units. All comparisons use the stored 121 output times and the true biochemical inputs to isolate integration from inverse search. This finite check concerns the frozen learned field, not the original coagulation simulator. The finer RK4 calculation is a numerical reference, not an exact solution; the check neither proves global convergence nor replaces retraining under a different solver.

\subsection{Biochemical-noise and timing details}
Tissue factor is unmeasured while fibrinogen and factors IX, VIII and V are
measured with independent Gaussian errors. Their standard deviations are
2\%, 5\% or 10\% of their respective training ranges. The same underlying
noise draws are used across severities and models. The four clot-size
observations remain exact, isolating biochemical measurement error from the
separate size-noise experiment. Measurements are not clipped. We compare
an exact-factor reference, noisy factors fixed to their measured values,
and the noise-aware proposal of Section~\ref{S-sec:noisy-assay-theory}.
The noise-aware method integrates all five inputs, four with informative assay likelihoods. All use 2,048 candidates,
temperature one and the same raw-size likelihood tolerance of 0.02.

Table~\ref{S-tab:noisy-factors} retains the pooled calibration rule: one
90th percentile of residual-to-standard-deviation ratios across calibration
cases and future times. These intervals differ from the per-time rank-46
intervals in the main comparison, but their point forecasts and parameter
estimates are identical.

\begin{table}[htbp]
\caption{Noisy biochemical factors with tissue factor unmeasured. Noise SD is a fraction of each training factor range, not a coefficient of variation. Each entry averages three split-specific summaries, with 50 calibration and 50 evaluation cases per split; the original checkpoint-selection cases are unused here. Forecast error and tissue-factor range-normalized MAE are percentages. Coverage, width and interval score concern pointwise 90\% size intervals; width and score use stored size units. Exact is a reference; fixed treats noisy factors as exact; aware integrates their assay uncertainty.}
\label{S-tab:noisy-factors}
\centering
\begin{tabular}{llrrrrr}
\toprule
Model / noise & Rule & Error & TF MAE & Cover & Width & Score \\
\midrule
SNODE / 0\% & Exact & 1.37 & 2.16 & 0.898 & 0.072 & 0.112 \\
SNODE / 2\% & Fixed & 1.68 & 2.60 & 0.890 & 0.097 & 0.139 \\
SNODE / 2\% & Aware & 1.63 & 2.58 & 0.899 & 0.083 & 0.117 \\
SNODE / 5\% & Fixed & 2.75 & 3.92 & 0.876 & 0.160 & 0.232 \\
SNODE / 5\% & Aware & 2.30 & 3.59 & 0.875 & 0.105 & 0.145 \\
SNODE / 10\% & Fixed & 4.96 & 6.59 & 0.893 & 0.303 & 0.417 \\
SNODE / 10\% & Aware & 2.84 & 4.75 & 0.909 & 0.134 & 0.190 \\
\midrule
MLP / 0\% & Exact & 1.21 & 1.60 & 0.850 & 0.059 & 0.112 \\
MLP / 2\% & Fixed & 1.56 & 2.08 & 0.862 & 0.090 & 0.148 \\
MLP / 2\% & Aware & 1.49 & 2.11 & 0.856 & 0.069 & 0.114 \\
MLP / 5\% & Fixed & 2.72 & 3.66 & 0.866 & 0.174 & 0.256 \\
MLP / 5\% & Aware & 2.18 & 3.32 & 0.848 & 0.092 & 0.147 \\
MLP / 10\% & Fixed & 5.05 & 6.53 & 0.888 & 0.354 & 0.482 \\
MLP / 10\% & Aware & 2.60 & 4.58 & 0.843 & 0.110 & 0.178 \\
\bottomrule
\end{tabular}
\end{table}

At 10\% factor noise, noise-aware conditioning reduces pooled error
from 4.96\% to 2.84\% for SNODE and from 5.05\% to 2.60\% for the MLP;
interval scores decrease from 0.417 to 0.190 and from 0.482 to 0.178.

As a forward-model diagnostic, supplying all five true biochemical inputs
gives average pooled errors of 1.34\% for SNODE and 1.17\% for the MLP on
the same evaluation cases. SNODE uses the conditional-prior latent mean
for this check, not the full-curve encoder or a marginalized latent prediction.

At 10\% noise, 15\% of measured factors lie outside the training bounds.
Among cases with all four measurements inside the box (31, 24 and 31
cases by split), errors decrease from 4.57\% to 3.01\% for SNODE and
from 4.59\% to 2.51\% for the MLP. This post-hoc subset does not separate
bounded-prior regularization from averaging over uncertain inputs.

At 10\% noise, noise-aware pointwise coverage is 90.9\% for SNODE but
84.3\% for MLP; whole-curve coverage is 71.3\% and 70.0\%.
Records include nominal levels 50\%, 70\%, 90\%, 95\%, raw coverage,
per-case errors and effective sample sizes.

After warm-up, five sequential single-case timings per condition include
candidate generation, prediction and weighting with 2,048 candidates and
121 times. Median latencies span 0.042--0.045 seconds for SNODE and
0.0012--0.0027 seconds for MLP, excluding loading, training, calibration
and acquisition. Batched throughput and peak tensor allocation are separate
quantities, not total system memory or a simulator speed-up. MLP selection
and refitting took 3.83--11.17 seconds; SNODE checkpoint loading took
0.006--0.121 seconds, not a retraining time.

\section{Sampled derivative and diagnostic sensitivity checks}

The POD--RBF surrogate is differentiated at 256 fixed Sobol points in each fitting-range box. Columns use input-range scaling; observation and forecast outputs use RMS norms. The sampled spectral-norm maxima below are properties of the surrogate, not uniform Lipschitz constants of the reference simulator. Validation maxima are observed errors, not global approximation bounds. Neither can be substituted into the finite-candidate theorem without additional uniform and reference-map control.

\begin{table}[htbp]\centering\caption{Empirical POD--RBF quantities at finite-difference step $10^{-3}$ of each input range. All error and derivative quantities use stored size units.}\begin{tabular}{rrrrr}\toprule Split & Early Jacobian max. & Future Jacobian max. & Early error max. & Future error max. \\\midrule

42 & 6.446 & 8.947 & 0.213 & 0.263 \\

43 & 6.279 & 8.248 & 0.172 & 0.248 \\

44 & 5.691 & 6.676 & 0.242 & 0.361 \\

\bottomrule\end{tabular}\end{table}

Reducing the step to $10^{-4}$ changes the displayed Jacobian maxima by less than $3\times10^{-5}$ in absolute value. This is a step-consistency check at sampled points, not an upper bound between them. The local-gain diagnostic instead divides all output derivatives by the fixed size tolerance 0.02; this positive scaling leaves AUROC unchanged but changes the absolute gain units.

The local future-gain diagnostic is also evaluated at all nine combinations of relative step and rank threshold in $\{10^{-4},10^{-3},10^{-2}\}$. Across these configurations and the three splits, AUROC ranges from 0.665 to 0.958. Full casewise derivatives and rankings are retained.

Two interval calculations are retained: a single pooled 90th percentile of calibration residual-to-SD ratios and the common per-time rank-46 rule. All headline comparisons use the latter. They share the same forecasts, diagnostic scores, selected tolerances and fitting/calibration/evaluation memberships. Differences between their intervals therefore concern the band rule rather than a change of predictor.

\bibliographystyle{plainnat}
\bibliography{references_amm,references_matcom_additions}